\documentclass[acmsmall, nonacm, screen]{acmart}

\AtBeginDocument{%
  \providecommand\BibTeX{{%
    \normalfont B\kern-0.5em{\scshape i\kern-0.25em b}\kern-0.8em\TeX}}}

\usepackage{pdfpages}
\usepackage{subcaption}
\usepackage{longtable}
\usepackage{tabularx}
\usepackage{appendix}

\newcounter{appendix}

\newcommand{\appendixtitle}{}
\newcommand{\appendixtitleFull}{%
  \space\Alph{appendix}.\quad\appendixtitle}
\AtBeginEnvironment{appendices}{%
  \captionsetup{labelsep=period}
  \counterwithin{figure}{appendix}
  \counterwithin{table}{appendix}
  \counterwithin{equation}{appendix}
  \counterwithin{section}{appendix}  
  \preto{\section}{%
    \clearpage
    \refstepcounter{appendix}
    \phantomsection 
    \addcontentsline{toc}{section}{\appendixtitleFull}%
  }
}

\makeatletter
\let\@authorsaddresses\@empty
\makeatother

\begin{document}

\title[Biomedical image analysis competitions: The state of current participation practice]{Biomedical image analysis competitions:\newline The state of current participation practice}

\author{Matthias Eisenmann}
\email{m.eisenmann@dkz-heidelberg.de}
\affiliation{%
 \institution{German Cancer Research Center (DKFZ) Heidelberg, Division of Intelligent Medical Systems}
 \country{Germany}
}

\author{Annika Reinke}
\affiliation{%
 \institution{German Cancer Research Center (DKFZ) Heidelberg, Division of Intelligent Medical Systems}
 \country{Germany}
}
\affiliation{%
 \institution{German Cancer Research Center (DKFZ) Heidelberg, HI Helmholtz Imaging}
 \country{Germany}
}
\affiliation{%
 \institution{Faculty of Mathematics and Computer Science, Heidelberg University, Heidelberg}
 \country{Germany}
}

\author{Vivienn Weru}
\affiliation{%
 \institution{German Cancer Research Center (DKFZ) Heidelberg, Division of Biostatistics}
 \country{Germany}
}

\author{Minu Dietlinde Tizabi}
\affiliation{%
 \institution{German Cancer Research Center (DKFZ) Heidelberg, Division of Intelligent Medical Systems}
 \country{Germany}
}

\author{Fabian Isensee}
\affiliation{%
 \institution{German Cancer Research Center (DKFZ) Heidelberg, Division of Medical Image Computing}
 \country{Germany}
}
\affiliation{%
 \institution{German Cancer Research Center (DKFZ) Heidelberg, HI Applied Vision Lab}
 \country{Germany}
}

\author{Tim J. Adler}
\affiliation{%
 \institution{German Cancer Research Center (DKFZ) Heidelberg, Division of Intelligent Medical Systems}
 \country{Germany}
}

\author{Patrick Godau}
\affiliation{%
 \institution{German Cancer Research Center (DKFZ) Heidelberg, Division of Intelligent Medical Systems}
 \country{Germany}
}

\author{Veronika Cheplygina}
\affiliation{%
 \institution{IT University of Copenhagen, Copenhagen}
 \country{Denmark}
}

\author{Michal Kozubek}
\affiliation{%
 \institution{Centre for Biomedical Image Analysis, Masaryk University, Brno}
 \country{Czech Republic}
}

\author{IEEE ISBI challenge organizers}
\authornote{Names and affiliations provided in Appendix~\ref{app:authors_organizers_isbi}.}

\author{MICCAI challenge organizers}
\authornote{Names and affiliations provided in Appendix~\ref{app:authors_organizers_miccai}.}

\author{IEEE ISBI challenge participants}
\authornote{Names and affiliations provided in Appendix~\ref{app:authors_participants_isbi}.}

\author{MICCAI challenge participants}
\authornote{Names and affiliations provided in Appendix~\ref{app:authors_participants_miccai}.}
\affiliation{%
 see Appendix
}

\author{Klaus Maier-Hein}
\affiliation{%
 \institution{German Cancer Research Center (DKFZ) Heidelberg, Division of Medical Image Computing}
 \country{Germany}
}

\author{Paul F. Jäger}
\affiliation{%
 \institution{German Cancer Research Center (DKFZ) Heidelberg, Interactive Machine Learning Group}
 \country{Germany}
}
\affiliation{%
 \institution{German Cancer Research Center (DKFZ) Heidelberg, HI Helmholtz Imaging}
 \country{Germany}
}

\author{Annette Kopp-Schneider}
\affiliation{%
 \institution{German Cancer Research Center (DKFZ) Heidelberg, Division of Biostatistics}
 \country{Germany}
}

\author{Lena Maier-Hein}
\email{l.maier-hein@dkfz-heidelberg.de}
\affiliation{%
 \institution{German Cancer Research Center (DKFZ) Heidelberg, Division of Intelligent Medical Systems}
 \country{Germany}
}
\affiliation{%
 \institution{German Cancer Research Center (DKFZ) Heidelberg, HI Helmholtz Imaging}
 \country{Germany}
}
\affiliation{%
 \institution{Faculty of Mathematics and Computer Science and Medical Faculty, Heidelberg University, Heidelberg}
 \country{Germany}
}
\affiliation{%
 \institution{National  Center  for  Tumor  Diseases  (NCT), NCT Heidelberg, a partnership between DKFZ and University Hospital Heidelberg}
 \country{Germany}
}

\renewcommand{\shortauthors}{Eisenmann et al.}

\begin{abstract}
\section*{Abstract}
The number of international benchmarking competitions is steadily increasing in various fields of machine learning (ML) research and practice. So far, however, little is known about the common practice as well as bottlenecks faced by the community in tackling the research questions posed. To shed light on the status quo of algorithm development in the specific field of biomedical imaging analysis, we designed an international survey that was issued to all participants of challenges conducted in conjunction with the IEEE ISBI 2021 and MICCAI 2021 conferences (80 competitions in total). The survey covered participants' expertise and working environments, their chosen strategies, as well as algorithm characteristics. A median of 72\% challenge participants took part in the survey. According to our results, knowledge exchange was the primary incentive (70\%) for participation, while the reception of prize money played only a minor role (16\%). While a median of 80 working hours was spent on method development, a large portion of participants stated that they did not have enough time for method development (32\%). 25\% perceived the infrastructure to be a bottleneck. Overall, 94\% of all solutions were deep learning-based. Of these, 84\% were based on standard architectures. 43\% of the respondents reported that the data samples (e.g., images) were too large to be processed at once. This was most commonly addressed by patch-based training (69\%), downsampling (37\%), and solving 3D analysis tasks as a series of 2D tasks. K-fold cross-validation on the training set was performed by only 37\% of the participants and only 50\% of the participants performed ensembling based on multiple identical models (61\%) or heterogeneous models (39\%). 48\% of the respondents applied postprocessing steps.
\end{abstract}


\begin{teaserfigure}
  \includegraphics[width=\textwidth]{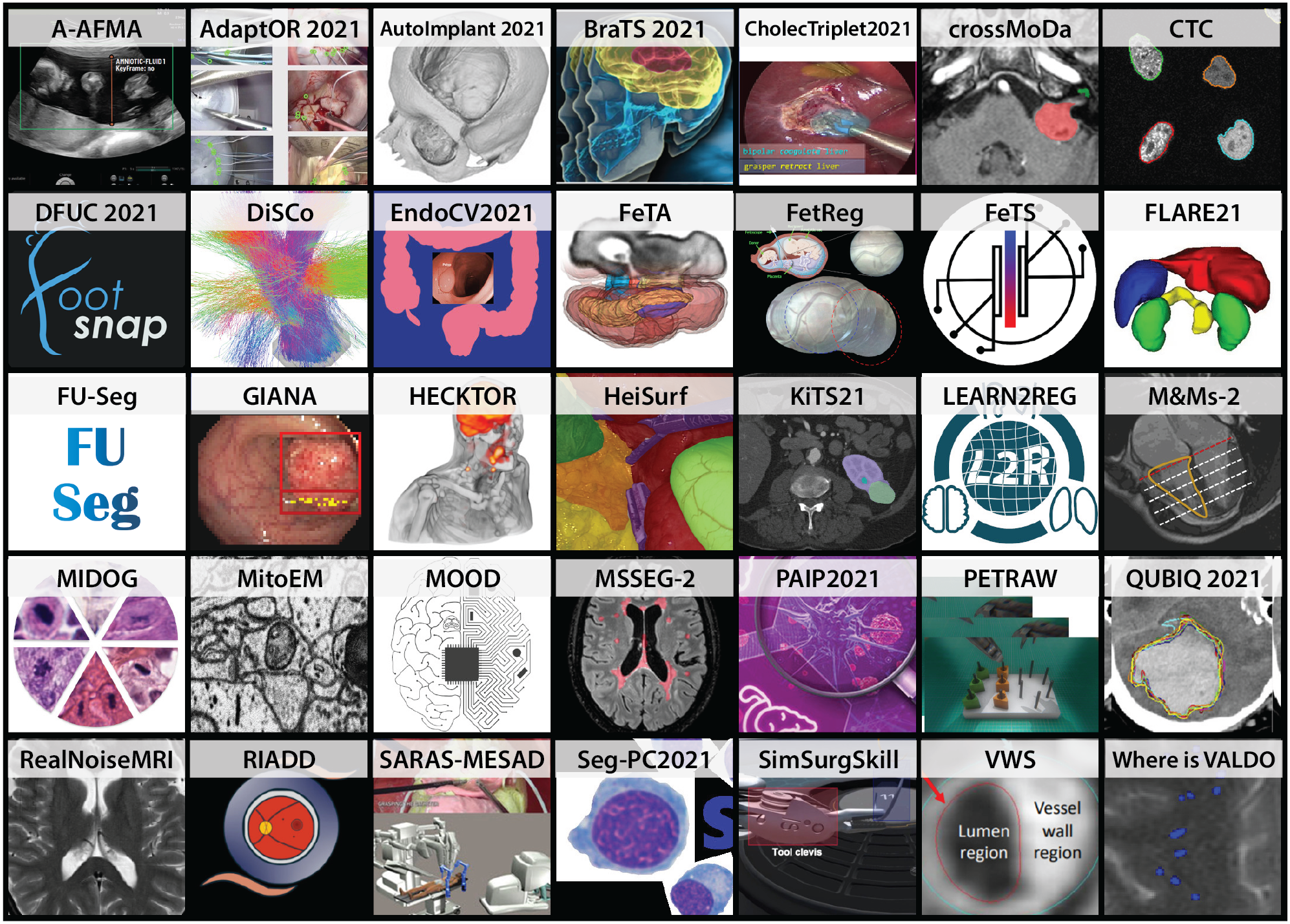}
  \caption{Overview of the 35 IEEE ISBI 2021 and MICCAI 2021 challenges in the scope of which 80 competitions with dedicated leaderboards were hosted, as detailed in App.~\ref{app:overview}. A representative image is shown for each challenge (labeled with its acronym, ordered alphabetically) contained in this meta-analysis. The number of competitions (tasks) with dedicated leaderboards varied between 1 and 21.}
  \Description{}
  \label{fig:teaser}
\end{teaserfigure}


\maketitle

\section{Purpose}
\label{sec:purpose}
Validation of biomedical image analysis algorithms is typically conducted through so-called challenges - large international benchmarking competitions that compare algorithm performance on identical datasets. Recent years have not only seen an increase in the complexity of the machine learning (ML) models used to solve the tasks, but also a tremendous increase of the scientific impact of challenges, with results often being published in prestigious journals (e.g., \cite{sage2015quantitative, chenouard2014objective, menze2014multimodal, ulman2017objective, maier2017challenge}) and the winner sometimes receiving important attention in terms of citations and monetary compensation. However, despite this impact, we identified a notable gap in the literature regarding insights into current common practice in challenges. To address this issue, we designed an international survey that was issued to all participants of challenges conducted in conjunction with the IEEE International Symposium on Biomedical Imaging (ISBI) and the International Conference on Medical Image Computing and Computer Assisted Intervention (MICCAI) in the year 2021 (80 leaderboards in total). This white paper presents the survey design and the summary of responses.

\section{Methods}
\label{sec:methods}
According to the BIAS guideline on biomedical challenges (established within the EQUATOR initiative)~\cite{maier-hein_bias_2020}, a biomedical image analysis challenge is defined as an “open competition on a specific scientific problem in the field of biomedical image analysis. A challenge may encompass multiple competitions related to multiple \emph{tasks}, whose participating teams may differ and for which separate rankings/leaderboards/results are generated”. As the term \emph{challenge task} is uncommon in the machine learning community, we will use the term \emph{competition} instead. The term \emph{challenge} will be reserved for the collection of competitions that are conducted under the umbrella of one dedicated organizational team/entity, represented by an acronym (Fig.~\ref{fig:teaser}).

The survey was developed in collaboration between Helmholtz Imaging and the Special Interest Group on biomedical image analysis challenges (SIG for Challenges) of the MICCAI society. It was structured in five parts and covered (1) general information on the team and the tackled task(s), (2) expertise and environment, (3) strategy for the challenge, (4) algorithm characteristics, and (5) miscellaneous information. Out of a maximum of 168 questions, the survey only showed questions that were relevant to the specific situation.

The organizers of all IEEE ISBI 2021 challenges (30 competitions across 6 challenges \cite{noauthor_ctc_nodate, noauthor_mitoem_nodate, ali2022assessing, noauthor_riadd_nodate, noauthor_segpc_2021_nodate, noauthor_a_afma_nodate}) and MICCAI 2021 challenges (50 competitions across 29 challenges \cite{nicholas_heller_2020_4674397, melanie_ganz_2021_4572640, reuben_dorent_2021_4573119, sandy_engelhardt_2021_4646979, moi_hoon_yap_2020_4646982, stefanie_speidel_2021_4572973, gabriel_girard_2021_4733450, jun_ma_2021_4596561, spyridon_bakas_2021_4573128, kelly_payette_2021_4573144, vincent_andrearczyk_2021_4573155, mattias_heinrich_2021_4573968, jens_petersen_2021_4573948, marc_aubreville_2021_4573978, carlos_martin_isla_2021_4573984, bjoern_menze_2021_4575204, spyridon_bakas_2021_4575162, fabio_cuzzolin_2021_4575197, jianning_li_2021_4577269, carole_sudre_2020_4600654, chun_yuan_2021_4575301, chuanbo_wang_2021_4575314, frederic_cervenansky_2021_4575409, jinwook_choi_2021_4575424, nwoye2022cholectriplet2021}) were invited to participate in the initiative and to bring us into contact with participants (if allowed by the challenge privacy policy) or distribute the survey link to them. The organizers were informed that the survey was targeted to those participants who submitted their solutions and would appear in the rankings. We created an individual survey website for each challenge to be able to accommodate the individual challenge schedule (i.e., challenge submission deadline). To avoid bias in survey responses, challenge participants were asked to complete the survey before knowing their position in the final ranking. The responses and feedback from the ISBI 2021 respondents were used to refine the survey for MICCAI, and are thus not included in the final results.

Where organizers were allowed to share the participants’ contact information (20 challenges), the survey was conducted in closed-access mode, meaning that the participants received individual links to the survey and reminders (if necessary). Fifteen surveys were executed in open-access mode, meaning that the organizers shared the link to the respective survey and took care of sending reminders. In these cases, we were not informed about the total number of challenge participants. Participants were asked to fill in one survey form for each final challenge submission (one per team if applicable).

\section{Results}
\label{sec:results}
Based on the positive responses by the organizers, 100\% (n = 80, see overview in App.~\ref{app:overview}) of all MICCAI (n = 50) and ISBI (n = 30) competitions were included in this study. These covered a wide range of problems related to semantic segmentation, image-level classification, registration, tracking, object detection, pipeline evaluation etc. (see Fig.~\ref{fig:teaser} and Tab.~\ref{tab:overview:competitions}). Based on the challenge outcome, the challenge organizers considered 11\% of the problems addressed as solved (partially: 79\%, not at all: 8\%). A median (min/max) of 72\% (11\%/100\%) of the challenge participants took part in the survey (this number can only be provided for the closed-access surveys). Overall, we received a total number of 292 completed survey forms (ISBI: 32, MICCAI: 260). Of those, 249 met our inclusion criteria for the analysis (i.e., second version of the survey refined for MICCAI 2021, survey completed by a lead developer, no duplicate responses from the same team).

\begin{figure}
  \includegraphics[clip, trim = 60 50 15 50, width=\textwidth]{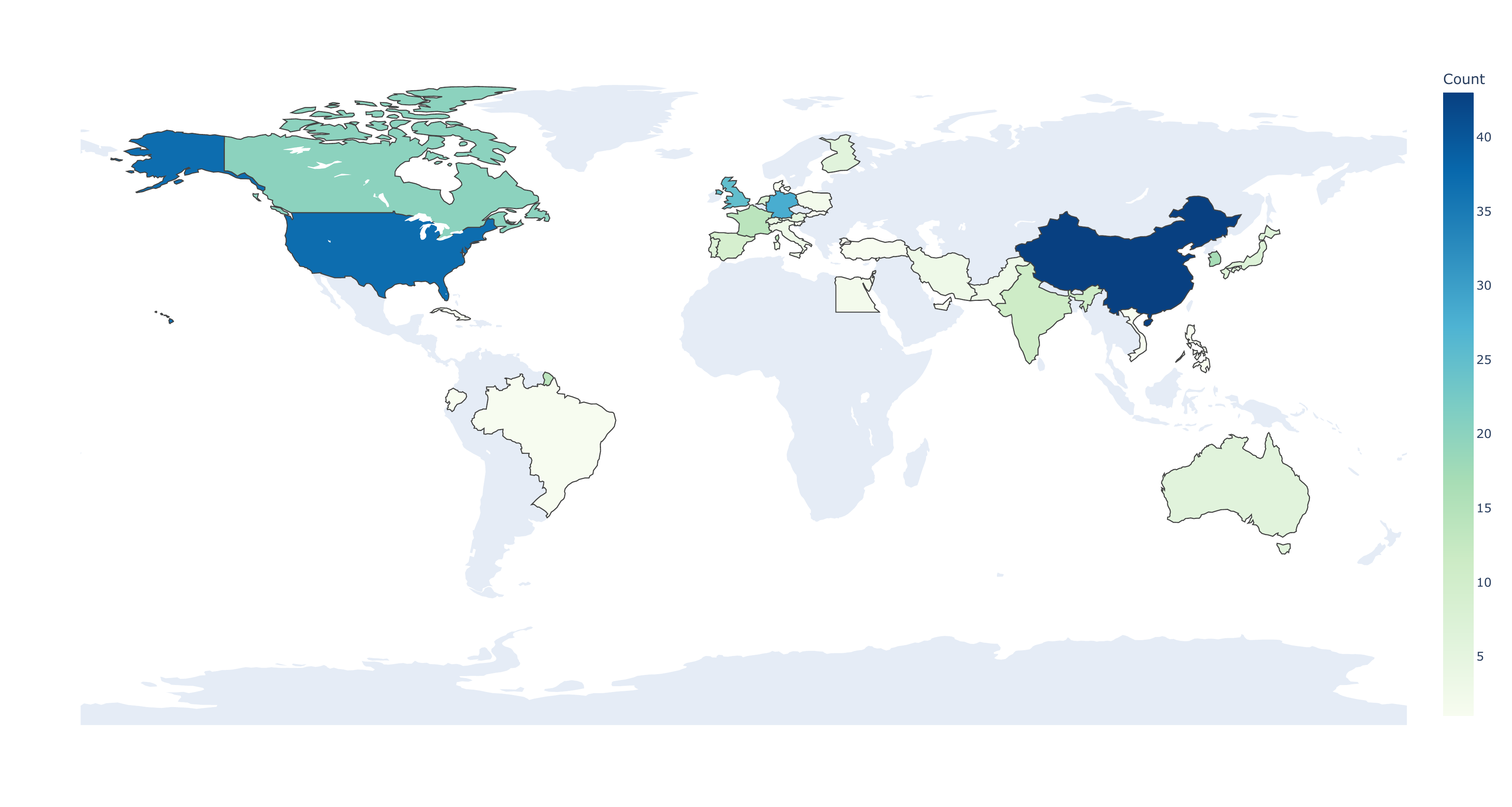}
  \caption{World map of challenge participants. The involved countries were extracted from the lead developers' and their supporting team members' affiliations. The participants originated from 34 different countries.}
  \label{fig:world_map_participants}
\end{figure}

The majority (86\%) of the challenge participants who responded were affiliated with academic institutions, 12\% were affiliated with industry, and 4\% did not belong to any institutions. The sum of these percentages exceeds 100\% because some respondents were affiliated with both academia and industry. A world map of involved countries based on the affiliations of the lead developers and their supporting team members is shown in Fig.~\ref{fig:world_map_participants}.

\begin{figure}
  \includegraphics[width=\textwidth]{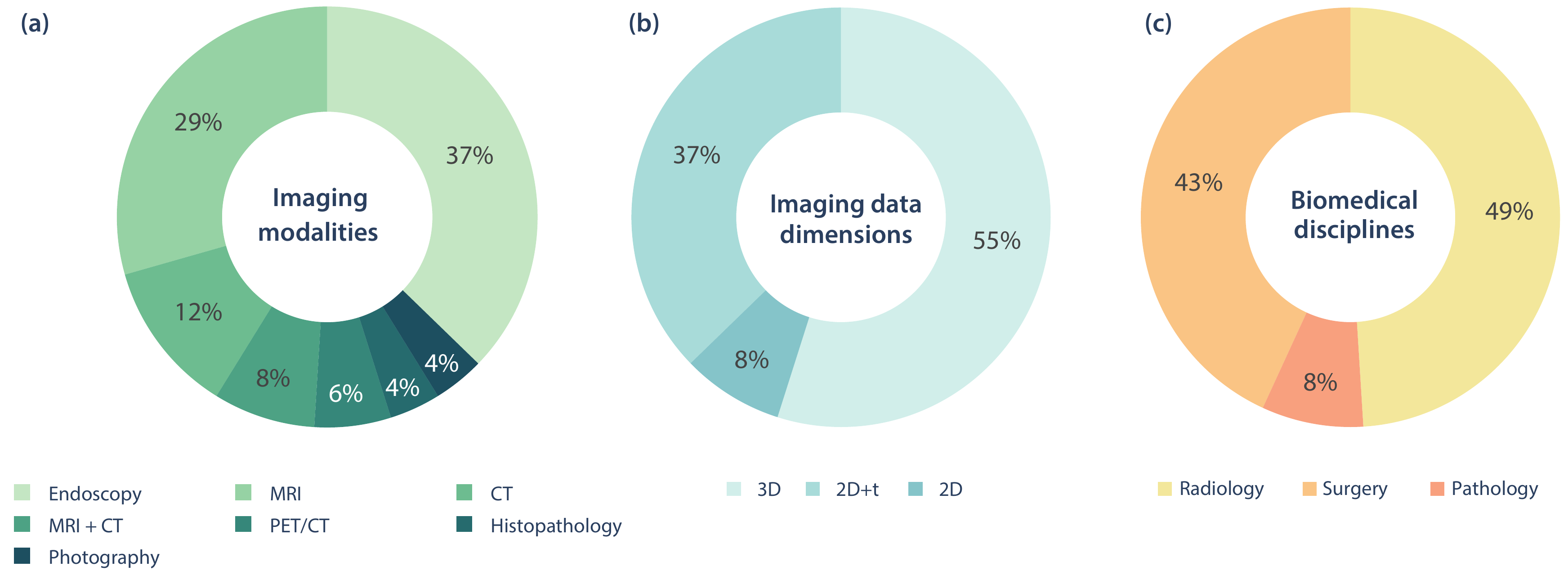}
  \caption{Overview of biomedical image analysis competitions. (a) Imaging modalities applied in the competitions, (b) dimensions of the datasets provided for the competitions, and (c) biomedical disciplines covered by the competitions.}
  \label{fig:competitions_characteristics}
\end{figure}

\begin{figure}
  \includegraphics[width=\textwidth]{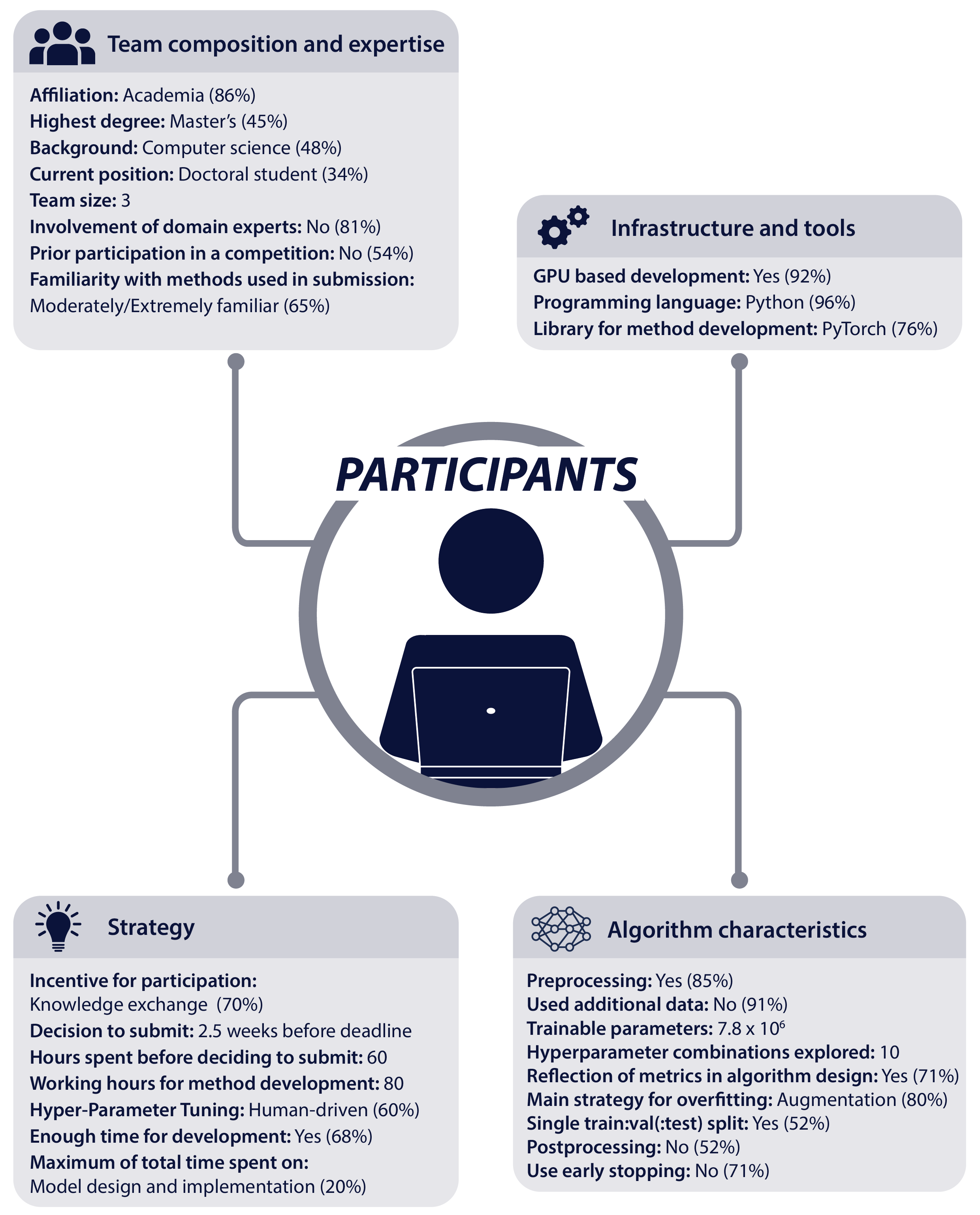}
  \caption{Profile of a “typical” competition participant. In case of categorical values, the majority vote of all participants was used. In case of continuous values, the median was taken. Algorithm characteristics are shown for DL-based approaches which were the majority of solutions (94\% of the respondents).}
  \label{fig:profile_participant}
\end{figure}

The profile of a “typical” competition participant is shown in Fig.~\ref{fig:profile_participant}. Further details are provided in the subsequent sections.

\subsection{Expertise and team composition}
Almost all respondents had an academic degree. In terms of the highest degree, 45\% had a master’s degree, 27\% had a doctoral degree, and 24\% had a bachelor’s degree. Their background was computer science (48\%), electrical engineering (17\%), or biomedical engineering (15\%). Mathematics, physics, and medicine were among the minority. When respondents were asked about their current position, 34\% answered that they were doctoral students, 19\% master's students, 10\% postdoctoral researchers, and 9\% professors. Also, 10\% were developers/engineers and 4\% were team leads or managers (4\%) in industry.

A median of 3 (Interquartile range (IQR)=[2, 4], min=1, max=11) team members contributed to the challenge submission. About a quarter (22\%) of the lead developers that worked in a team answered that they mainly worked alone. 16\% of all respondents stated that they participated entirely alone in the challenge. Interestingly, less than half of the respondents mentioned that they had a regular meeting with a supervisor (43\%) or colleagues/other method experts (47\%) to discuss their methods and/or results. In 12\% of the teams, multiple team members worked on/implemented a single approach together. Multiple team members explored and implemented diverse approaches simultaneously in 27\% of the teams.

It is noteworthy that only 22\% of the teams (n = 47) had a domain expert involved in their team. The experts contributed in various phases (multiple selections allowed): problem/data exploration (66\%), algorithm design (including pre- and postprocessing) (53\%), failure analysis (23\%), and/or tuning and optimization (9\%).

More than half of the respondents (54\%) did not have experience in participating in a machine learning competition, while the remaining had participated in a median of 2 (IQR=[1, 5], min=1, max=14) challenges before. The most experienced member of each team had a median of 2 (IQR=[1, 4], min=1, max=22) challenge experiences.

\begin{figure}
  \includegraphics[width=\textwidth]{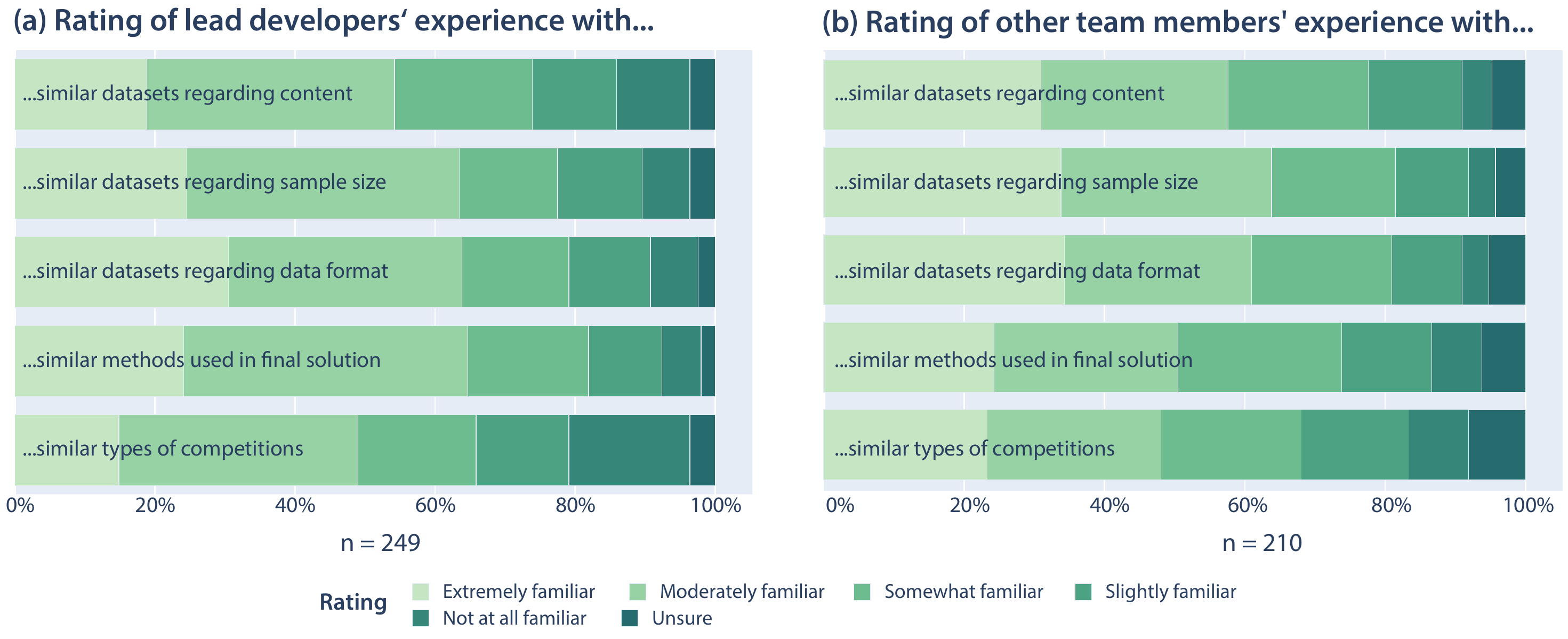}
  \caption{Rating of the experience with different aspects of the competition design. (a) Lead developers were asked to rate their own experience. (b) Those lead developers who worked in a team rated the experience of their team members as well.}
  \label{fig:experience_rating}
\end{figure}

On a Likert scale, 49\% of the respondents rated their own experience with similar types of competitions as moderately or extremely familiar (Fig.~\ref{fig:experience_rating}a). Regarding similar methods used in the final solution, almost two-thirds of the respondents felt moderately/extremely familiar (65\%), whereas 16\% felt slightly/not at all familiar. The percentage of respondents that rated their experience with similar datasets regarding data format, sample size, and content as moderately/extremely familiar was 64\%, 64\%, and 54\%, respectively (slightly/not at all familiar: 18\%, 26\%, and 30\%).

Overall, 48\% of the respondents who worked in a team (n = 210) rated the experience of their other team members as moderately/extremely familiar regarding similar types of competitions (Fig.~\ref{fig:experience_rating}b). Regarding similar methods used in the final solution, 50\% of the respondents rated their team members moderately/extremely familiar. The percentage of respondents that rated their experience with similar datasets regarding data format, sample size, and content as moderately/extremely familiar was only 61\%, 64\%, and 58\%, respectively (slightly/not at all familiar: 14\%, 14\%, and 18\%).

\subsubsection{Compute infrastructure}
25\% of all respondents thought that their infrastructure was a bottleneck. The vast majority of respondents used graphics processing unit (GPU) resources (92\%). Of those, 16\% used a single GPU that had to be shared with others and 27\% used a single GPU that they had to themselves. Multiple GPUs in a workstation were accessible to 25\%. Not professionally managed GPU clusters were used by 18\%, whereas 24\% used a professionally managed GPU cluster.

The total training time of all models trained during method development including failure models (across the team if several models were developed simultaneously) was estimated to be a median of 267 (IQR=[50, 720], min=1, max=10,000) GPU hours. Regarding the final model only, the total training time was estimated to be a median of 24 (IQR=[6, 69], min=0.05, max=2,517) GPU hours.

\subsubsection{Software frameworks and tools}
For 72\% of the competitions, submission to the final testing stage based on Docker containers was offered. Submission via websites/platforms and via e-mail was possible in 12\% of the competitions, respectively, whereas 4\% of the competition required a competition-specific framework.

Python was the main programming language used for implementation of the respondents’ methods (96\%), followed by MATLAB (2\%).

\begin{figure}
     \centering
     \begin{subfigure}[b]{0.45\textwidth}
         \centering
         \includegraphics[width=\textwidth]{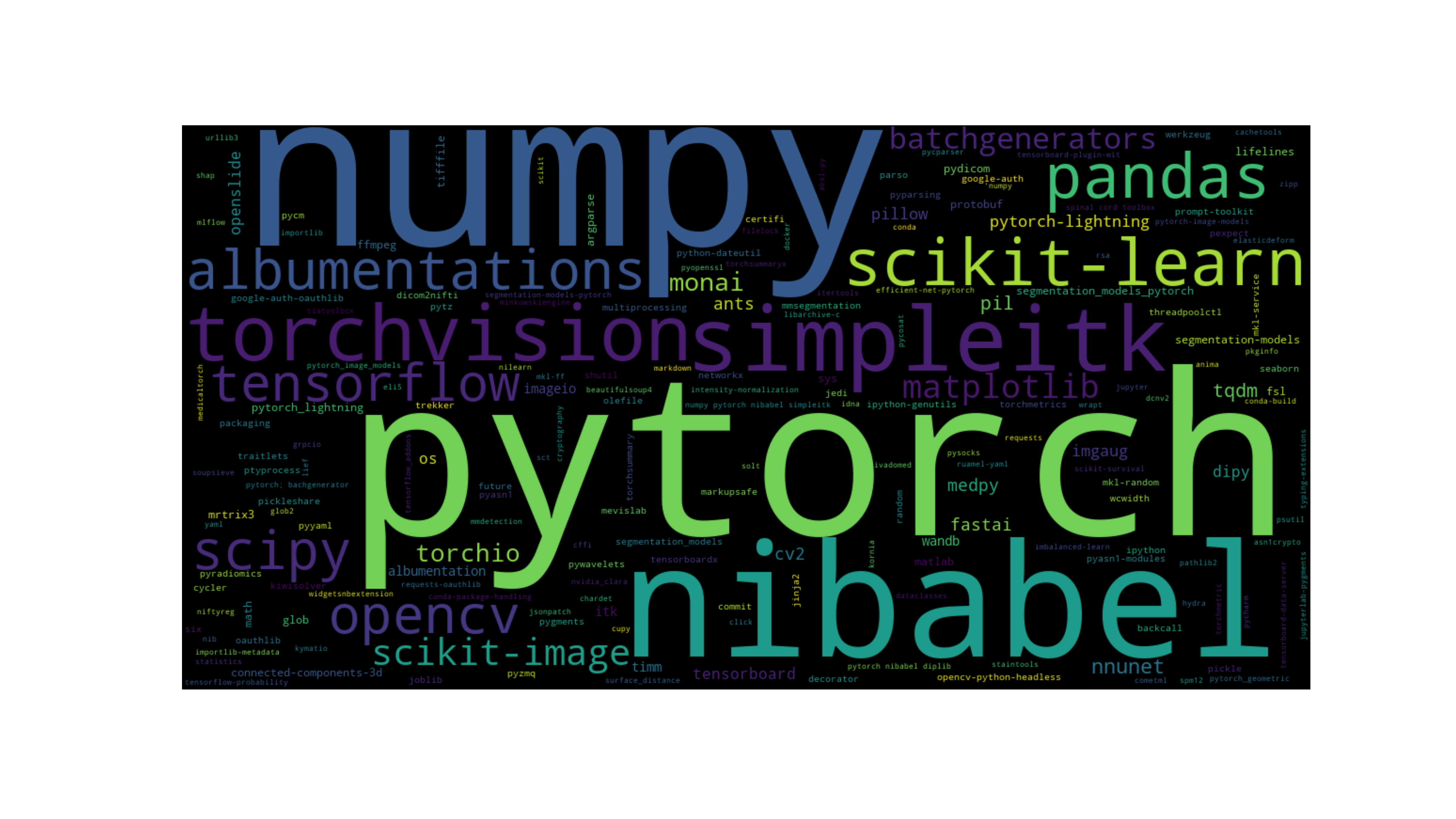}
         \caption{Low-level, core and high-level libraries}
         \label{fig:libraries_implementation}
     \end{subfigure}
     \hfill
     \begin{subfigure}[b]{0.45\textwidth}
         \centering
         \includegraphics[width=\textwidth]{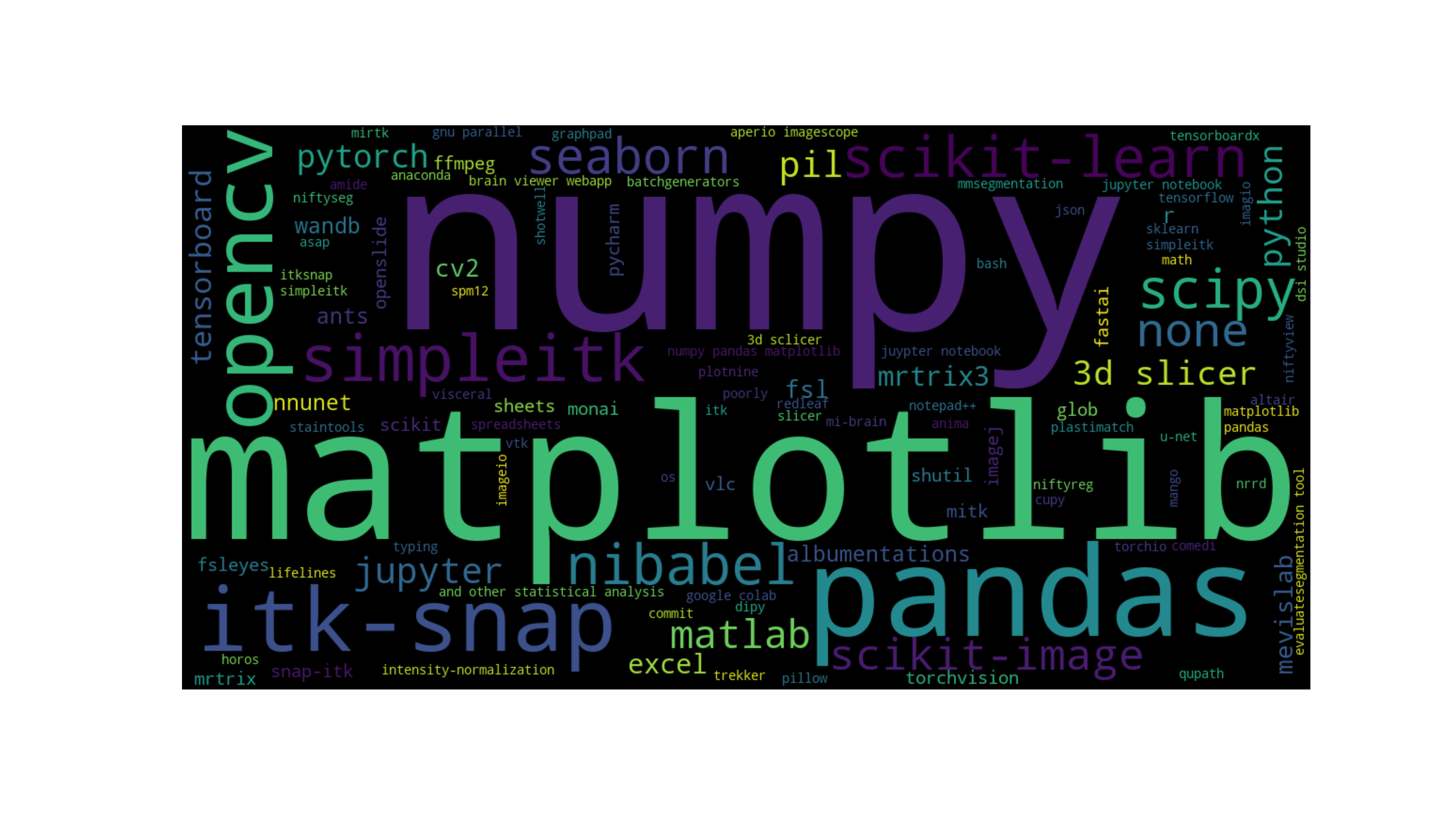}
         \caption{Tools used for analyzing data}
         \label{fig:tools_data_analysis}
     \end{subfigure}
     \hfill
     \begin{subfigure}[b]{0.45\textwidth}
         \centering
         \includegraphics[width=\textwidth]{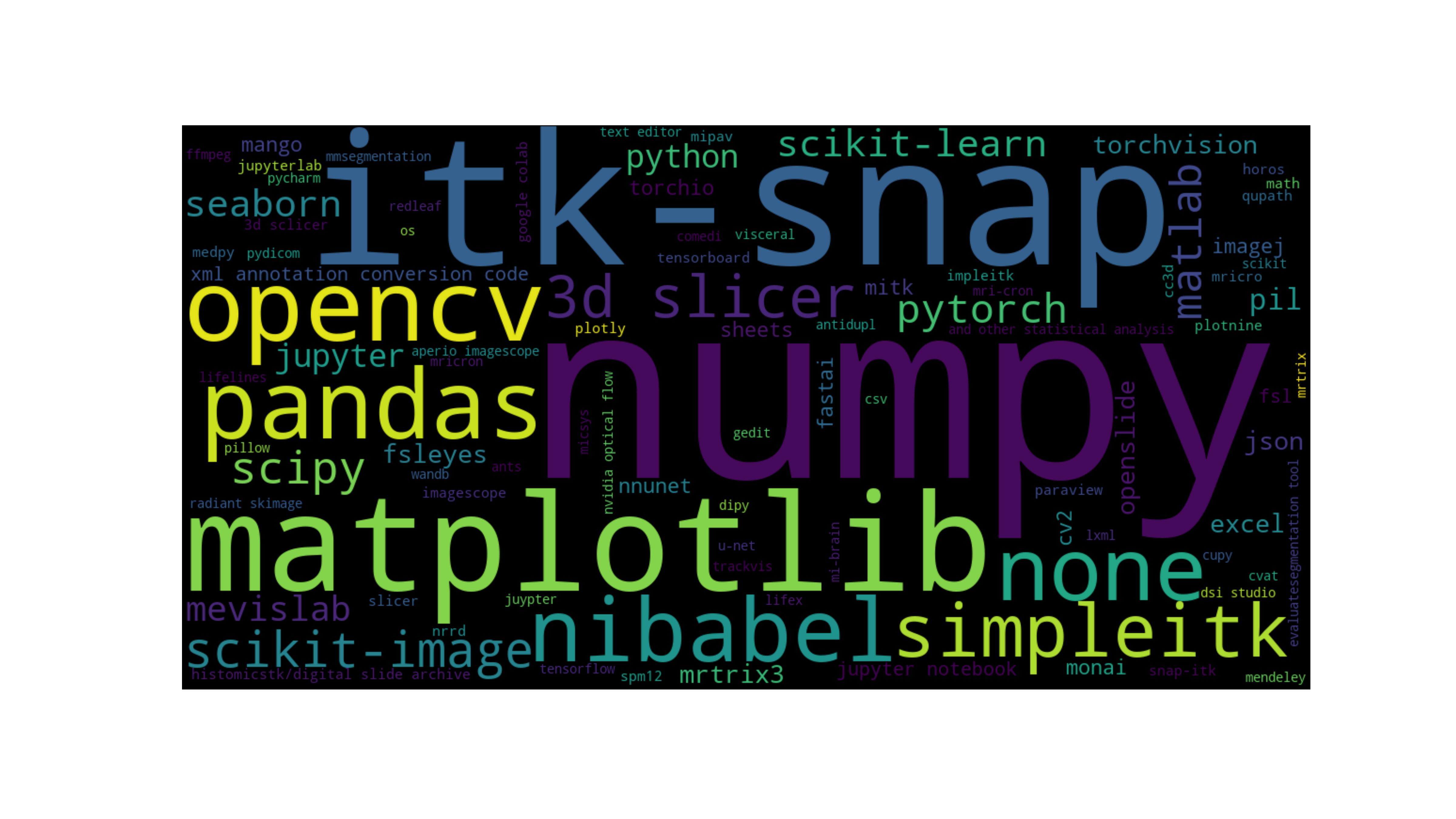}
         \caption{Tools used for analyzing reference data}
         \label{fig:tools_reference_data_analysis}
     \end{subfigure}
     \hfill
     \begin{subfigure}[b]{0.45\textwidth}
         \centering
         \includegraphics[width=\textwidth]{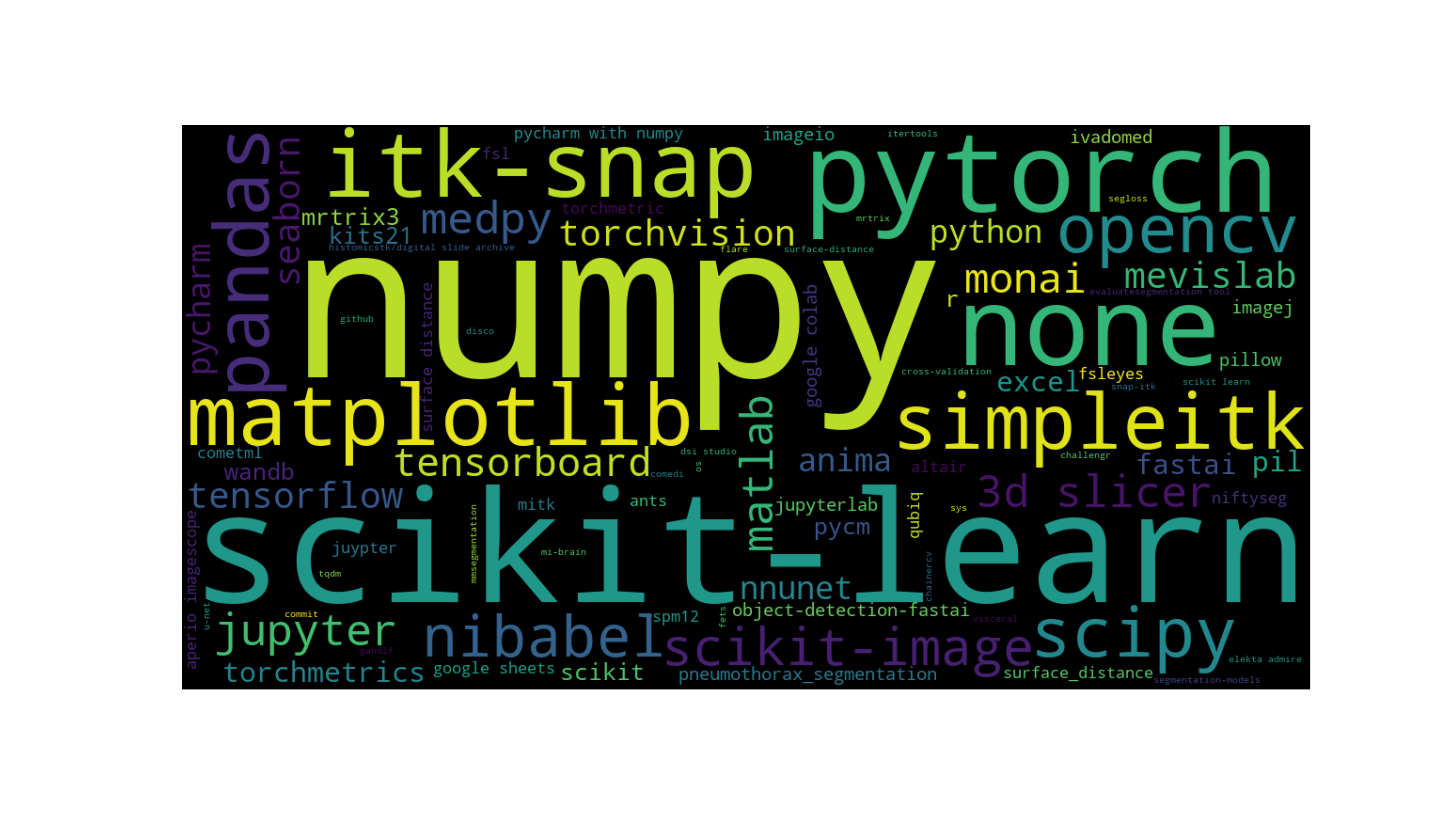}
         \caption{Tools used for internal evaluation}
         \label{fig:tools_internal_evaluation}
     \end{subfigure}
     \hfill
     \begin{subfigure}[b]{0.45\textwidth}
         \centering
         \includegraphics[width=\textwidth]{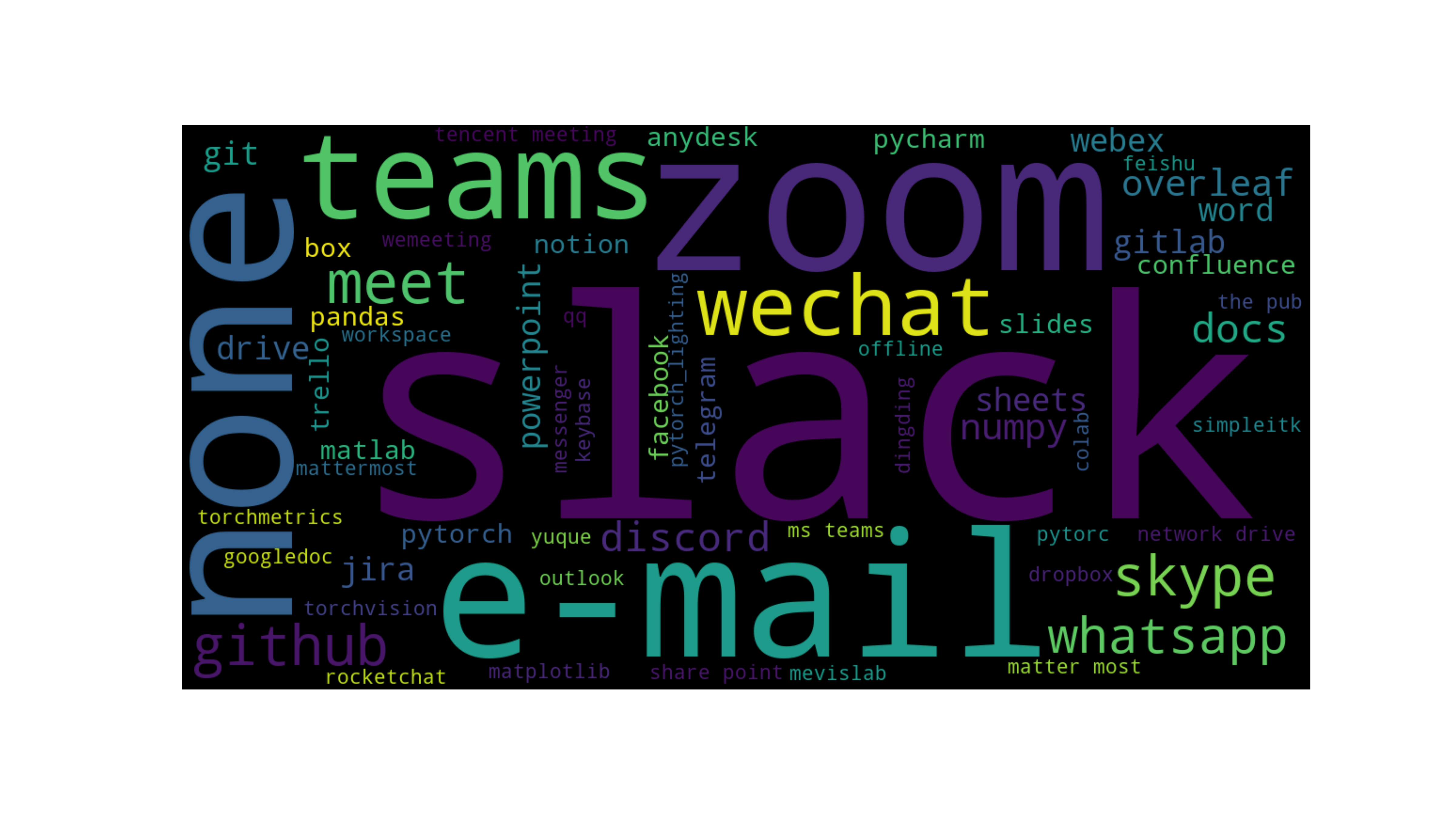}
         \caption{Tools used for team communication}
         \label{fig:tools_team_communication}
     \end{subfigure}
     \caption{Word cloud of (a) low-level, core, and high-level libraries used for implementation of method(s), (b) tools/frameworks/packages used for analyzing data, (c) tools/frameworks/packages for analyzing annotations/reference data, (d) tools/frameworks/packages used for internal evaluation, (e) tools/frameworks/packages used for team communication. The size of a term corresponds to the number of times it was mentioned in the survey. The free text responses were cleaned manually. Missing responses are encoded as "none".}
\end{figure}

A summary of the free text responses (cleaned manually) given for specific types of software used is provided in dedicated word clouds:
\begin{enumerate}
    \item The top low-level, core, and high-level libraries used for implementation of method(s) were PyTorch (76\%), NumPy (74\%), NiBabel (34\%), SimpleITK (33\%), and torchvision (29\%) (see Fig.~\ref{fig:libraries_implementation}).
    \item The most commonly used tools/frameworks/packages used for analyzing data were NumPy (37\%), Matplotlib (26\%), pandas (24\%), ITK-SNAP (15\%), and OpenCV (14\%) (see Fig.~\ref{fig:tools_data_analysis}).
    \item The most commonly used tools/frameworks/packages for analyzing annotations/reference data were NumPy (27\%), ITK-SNAP (23\%), Matplotlib (17\%), OpenCV (12\%), and pandas (11\%) (see Fig.~\ref{fig:tools_reference_data_analysis}).
    \item The most popular tools/frameworks/packages used for internal evaluation were NumPy (27\%), scikit-learn (15\%), PyTorch (15\%), ITK-SNAP (9\%), and Matplotlib (8\%) (see Fig.~\ref{fig:tools_internal_evaluation}).
    \item The most common tools/frameworks/packages used for team communication were Slack (20\%), Zoom (20\%), e-mail (15\%), Microsoft Teams (15\%), and WeChat (11\%) (see Fig.~\ref{fig:tools_team_communication}).
\end{enumerate}

\newpage
\subsection{Strategy for the challenge}
This part refers to all exploration aspects that led to the final solution.

\subsubsection{Decision to participate}
Knowledge exchange was the most important incentive for participation (mentioned by 70\%; respondents were allowed to pick multiple answers), followed by the possibility to compare their own method to others (65\%), having access to data (52\%), being part of an upcoming challenge publication (50\%), winning a challenge (42\%), and networking (31\%). The awards/prize money was important to only 16\% of the respondents. Note in this context that some competitions do not offer access to data or prize money. 

The respondents were also asked when they decided to submit results for the competition. They reported a median time of 2.5 (IQR=[1, 5], min=0.3, max=28) weeks prior to the submission deadline. \emph{Before deciding to submit} results, the respondents invested 60 working hours (median) (IQR=[24, 150], min=1, max=5,000) in the preparation.

Most of the work \emph{prior to the decision to submit} results was dedicated to method development (including preprocessing, model development and postprocessing) (73\%), running a baseline method (46\%), analyzing data and annotations (44\%), hyperparameter tuning (31\%), literature research (26\%), analysis of failure cases (18\%), and checking that the challenge design (e.g., metrics, ranking) is reasonable (15\%) (respondents were allowed to pick up to three answers).

\subsubsection{Method development}
When asked about their approach for method development, 42\% of respondents stated that they went through related literature and built upon/modified existing work. 25\% went through related literature and loosely based their approach on previous work. The reference method from the literature that was closest to the problem at hand was identified and reimplemented/optimized for the task at hand by 15\%. 9\% went through related literature, but did not find anything suitable, built something completely new and mostly unrelated to existing work. Interestingly, 4\% reported to have neither conducted any literature research nor obtained any references or reused any code, but worked based on their intuition alone.

\begin{figure}
  \includegraphics[clip, trim = 10 40 60 70, width=\textwidth]{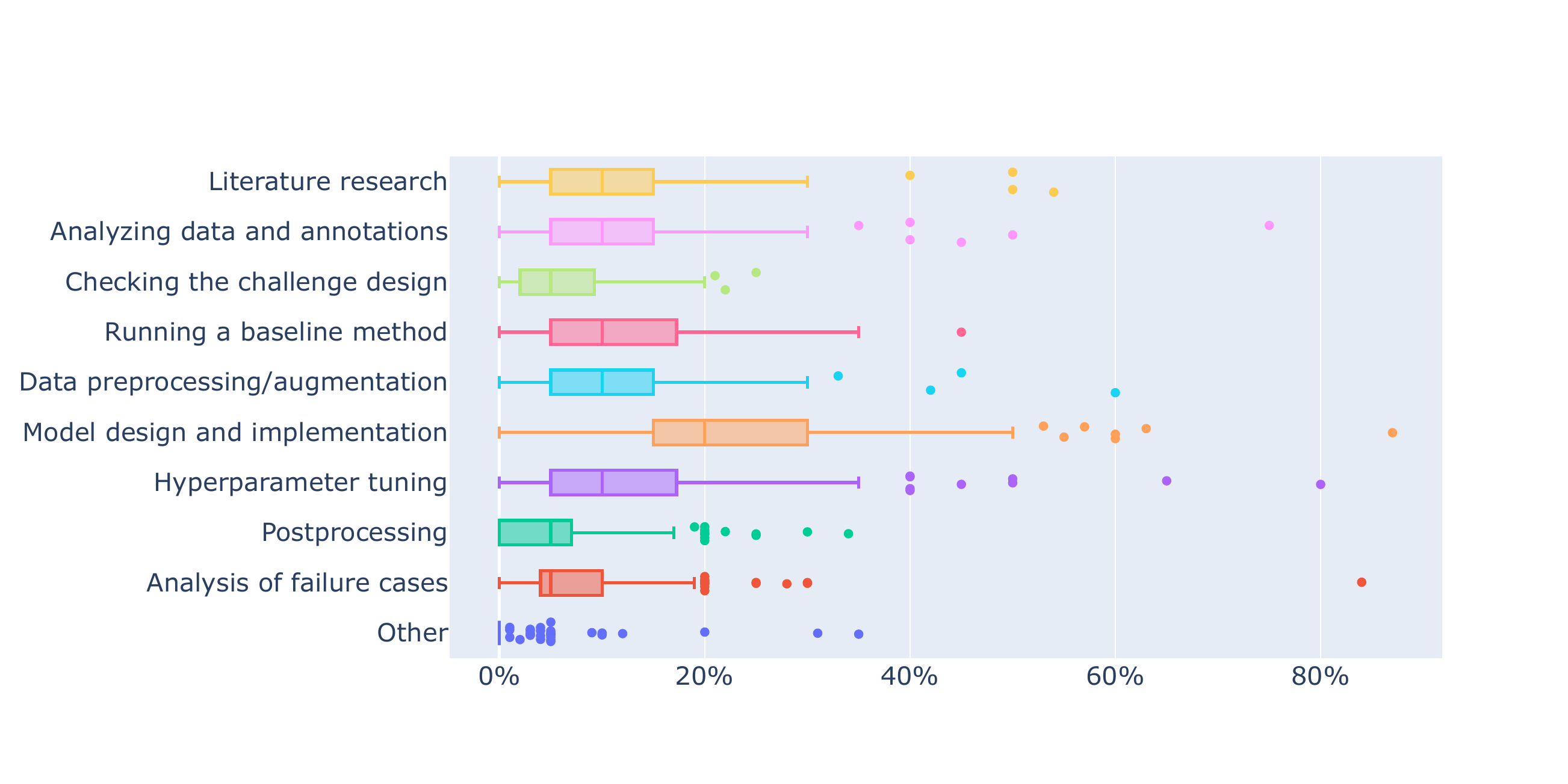}
  \caption{The respondents (n = 249) distributed 100\% of their total time spent on method development to the options given on the y-axis. The center line in the boxplots shows the median, the lower, and upper border of the box represent the first and third quartile. The whiskers extend to the lowest value still within 1.5 interquartile range (IQR) of the first quartile, and the highest value still within 1.5 IQR of the third quartile.}
  \label{fig:time_algo_dev_distribution}
\end{figure}

\begin{figure}
  \includegraphics[clip, trim = 45 140 100 100, width=\textwidth]{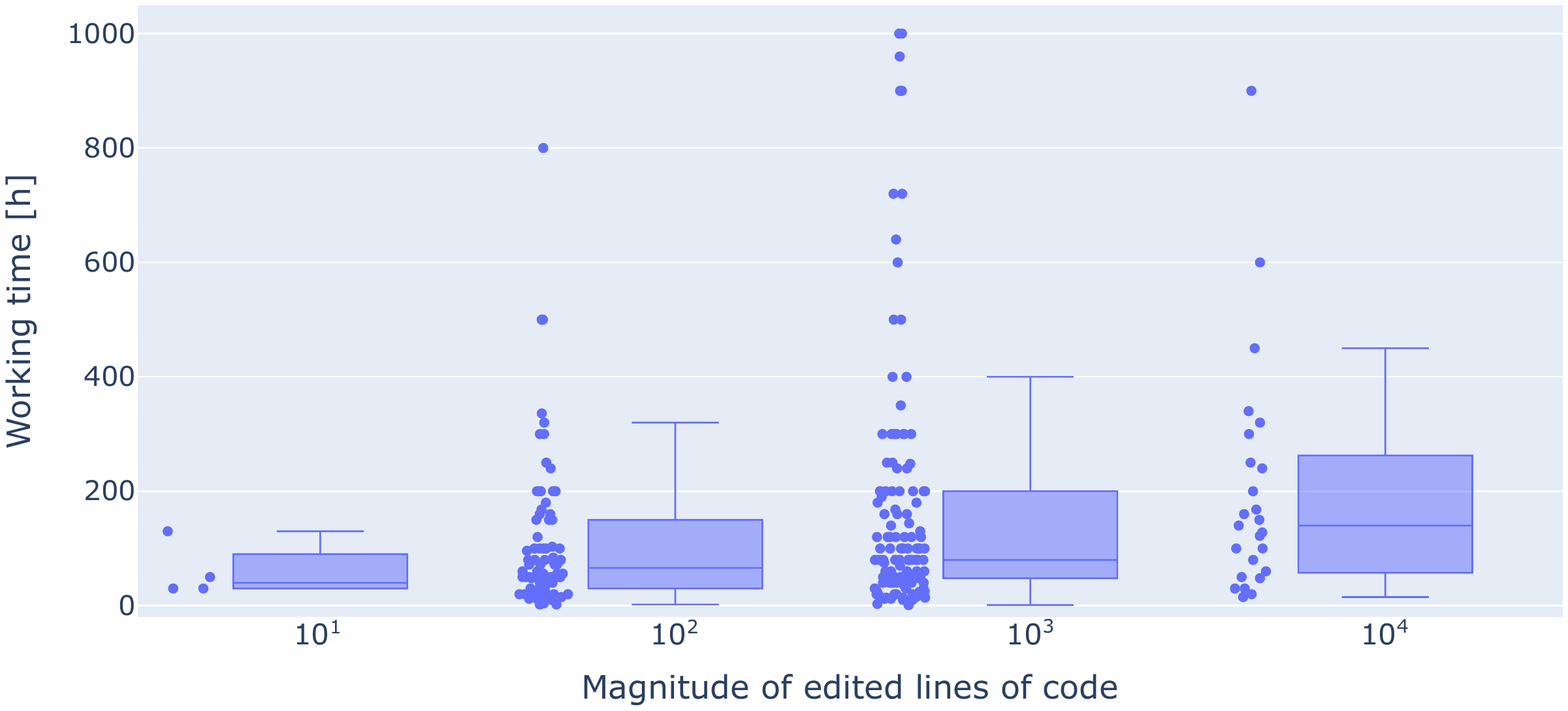}
  \caption{Relation of total time spent on method development to edited lines of code. One data point represents one of the 249 survey responses. The center line in the boxplots shows the median, the lower, and upper border of the box represent the first and third quartile. The whiskers extend to the lowest value still within 1.5 interquartile range (IQR) of the first quartile, and the highest value still within 1.5 IQR of the third quartile.}
  \label{fig:relation_working_time_editedLOC}
\end{figure}

Half of the respondents reimplemented a method based on a publication. A code base of the baseline method was used by 57\%. The number of edited lines of code of the final solution were reported  within the order of magnitudes $10^1$, $10^2$, $10^3$, and $10^4$ by 2\%, 33\%, 51\%, and 10\% of the respondents, respectively. A median of 80 (IQR=[42, 200], min=1, max=5,000) working hours was spent on method development in total. Most of their time was spent on model design and implementation (Fig.~\ref{fig:time_algo_dev_distribution}). The relation of the total working time spent on method development to the edited lines of code is shown in Fig.~\ref{fig:relation_working_time_editedLOC}.

We were also interested in the types of decisions that were made throughout the method development. The respondents reported more human-driven decisions, e.g. parameter setting based on expertise, than empirical decisions, e.g. automated hyperparameter tuning via grid search (human-driven: median=60\%, IQR=[40\%, 80\%], min=0\%, max=100\%; empirical: median=40\%, IQR=[10\%, 58\%], min=0\%, max=100\%).

The main focus of the survey was to cover characteristics of solutions based on deep learning. 94\% of the respondents used a deep learning-based approach (n = 233). For those approaches, most time was spent selecting one or multiple existing architectures (e.g., U-Net, ResNet, DenseNet) that best match the task (45\%) as well as configuring the data augmentation (33\%) (Tab.~\ref{tab:dl_aspects}, respondents were allowed to pick up to three answers).

\begin{longtable}{ p{10cm} p{3cm} }
\label{tab:dl_aspects} \\
\caption{Aspects that respondents who submitted a DL-based solution spent most time on. The respondents were allowed to pick up to three answers.} \\
\toprule
Aspect & Percentage of respondents (n = 233) \\* \midrule
\endfirsthead
\multicolumn{2}{c}%
{{\bfseries Table \thetable\ continued from previous page}} \\
\toprule
Aspect & Percentage of respondents (n = 233) \\* \midrule
\endhead
Selecting one or multiple existing architectures (e.g., U-Net, ResNet, DenseNet) that best match the task & 45\% \\
Configuring the data augmentation & 33\% \\
Configuring the template architecture (e.g., How deep? How many stages/pooling layers?) & 28\% \\
Exploring existing loss functions & 25\% \\
Ensembling & 22\% \\
Choosing a template architecture & 17\% \\
Other new methodological contributions (besides architecture; e.g., task-specific loss) & 16\% \\
Optimizing postprocessing (e.g., aggregating predictions during inference) & 15\% \\* \bottomrule
\end{longtable}

Among all respondents, 17\% \emph{explored} additional data (i.e. data not provided for the respective competition). In this context, “exploring” also means analyzing data that was incorporated in the final solution. Of those, public (48\%) and private (36\%) biomedical data for same type of task were mainly explored (Tab.~\ref{tab:data_exploration_vs_usage}, respondents were allowed to pick multiple answers).

\begin{longtable}{ p{7cm} p{3cm} p{3cm} }
\label{tab:data_exploration_vs_usage} \\
\caption{Comparison of different types of additional data (i.e. data not provided for the respective competition) that was explored by all respondents and used in the DL-based solutions. In this context, “exploring” also means analyzing data that was incorporated in the final solution. The respondents were allowed to pick multiple answers.} \\
\toprule
Type of data & Exploration of additional data (n = 42) & Usage of additional data in final solution (n = 20) \\* \midrule
\endfirsthead
\multicolumn{2}{c}%
{{\bfseries Table \thetable\ continued from previous page}} \\
\toprule
Type of data & Percentage of respondents (n = 42) exploring additional data & Percentage of respondents (n = 20) using additional data in final solution \\* \midrule
\endhead
Biomedical data for same type of task - public & 48\% & 40\% \\
Biomedical data for same type of task - private & 36\% & 25\% \\
Biomedical data for different type of task - public & 21\% & 15\% \\
Biomedical data for different type of task - private & 10\% & 0\% \\
Non-biomedical data - public & 7\% & 5\% \\
Non-biomedical data - private & 0\% & 0\% \\
Re-annotated data & 2\% & 0\% \\* \bottomrule
\end{longtable}

The survey revealed that almost one third of the respondents did not have enough time for development. A majority thereof (65\%) felt that more time in the scale of weeks would have been beneficial (months: 18\%, days: 14\%). 38\% definitely expected a substantial performance boost of their approach had they had more time (probably: 38\%, possibly: 20\%, possibly not: 4\%, unsure: 1\%).

\subsection{Algorithm characteristics}
This part refers to the final solutions based on deep learning (submitted results or algorithm).

\subsubsection{Data}
Among the deep learning-based approaches, 9\% \emph{used} additional data (i.e. data not provided for the respective challenge) in their final solution (n = 20). Of those, public (40\%) and private (25\%) biomedical data for same type of task were mainly used (Tab.~\ref{tab:data_exploration_vs_usage}, respondents were allowed to pick multiple answers). In 55\% of the cases, the additional data was used for pre-training, in 50\% for co-training.

\subsubsection{Network topology}
In total, 84\% of the networks were based on a commonly-used computer vision architecture (e.g., U-Net, ResNet, DenseNet). One third of all networks were pre-trained on another image dataset such as ImageNet. The networks had a median of 7.8E+06 (IQR=[1.9E+05, 3.1E+07], max=3.8E+08) trainable parameters. A median of 10 (IQR=[5, 20], max=1.10E+07) hyperparameter combinations were roughly explored. An architecture search to find the final network architecture was performed by 13\% of the respondents. The majority of respondents (80\%) approached the challenge with a matching architecture type (e.g., segmentation network for a segmentation task), whereas 5\% followed a non-standard approach (e.g., semantic segmentation network used for a detection task). The architecture was modified in a specific way to improve the performance on the challenge dataset by more than half of the respondents (57\%). Also, in 71\% of the cases, the metrics used to evaluate the challenge were taken into account while searching for hyperparameters. When asking about the strategy to avoid overfitting, data augmentation, batch normalization, dropout, and weight decay were mentioned by 80\%, 66\%, 44\%, and 43\%, respectively.

\subsubsection{Data augmentation}
\begin{figure}
  \includegraphics[width=0.7\textwidth]{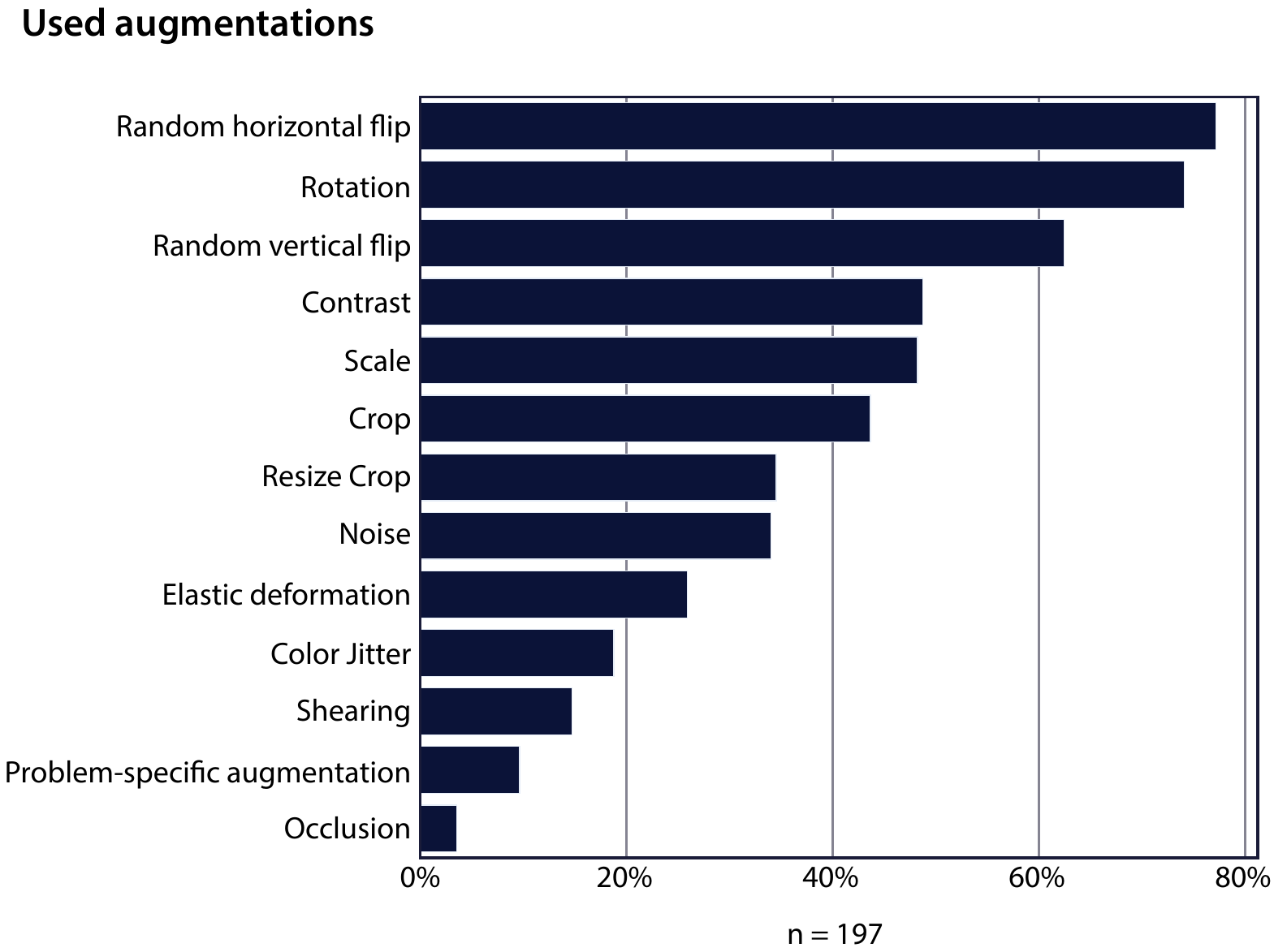}
  \caption{Data augmentations used by the respondents. The respondents were allowed to pick multiple answers.}
  \label{fig:data_augmentations}
\end{figure}

85\% of the respondents applied data augmentation (n = 197). Random horizontal flip (77\%), rotation (74\%), and random vertical flip (62\%) were the most frequently used augmentations (Fig.~\ref{fig:data_augmentations}).

43\% of the respondents reported that the data samples were too large to be processed at once (for example due to GPU memory constraints). Of those, 69\% worked with 3D data. This issue was mainly solved by patch-based training (cropping) (69\%), downsampling to a lower resolution (37\%), solving 3D analysis task as a series of 2D analysis tasks (per z-slice approach) and some postprocessing (18\%), and solving time-lapse analysis task as a series of single-frame analysis tasks (per time-frame approach) and some postprocessing (5\%) (respondents were allowed to pick multiple answers).

\subsubsection{Optimization}
The respondents mainly optimized Cross-Entropy Loss (39\%), Combined CE and Dice Loss (32\%), Dice Loss (26\%), custom-designed loss functions for problem (9\%), and Mean Squared Error Loss (5\%) (respondents were allowed to pick multiple answers). In the free-text responses (19\%), Focal Loss and Binary Cross Entropy Loss were mentioned most frequently.

29\% of the respondents used early stopping, 12\% used warmup. Internal evaluation via a single train:val(:test) split was performed by more than half of the respondents (52\%). K-fold cross-validation on the training set was performed by 37\%. 6\% did not perform any internal evaluation.

\subsubsection{Ensemble methods}
The final solution of half of the respondents was a single model trained on all available data. An ensemble of multiple identical models, each trained on the full training set but with a different initialization (random seed), was proposed by 6\%. 21\% proposed an ensemble of multiple identical models, each trained on a randomly drawn subset of the training set (regardless of whether the same seed was used or not). 9\% reported having ensembled multiple different models and trained each on the whole training set (different seeds). 8\% ensembled multiple different models, each trained on a randomly drawn subset of the training set (regardless of whether the same seed was used or not). If multiple models were used, the final solution was composed of a median of 5 (IQR=[3, 6], max=21) models. 48\% of the respondents applied postprocessing steps.

\section{Conclusion}
In this manuscript, we presented the results of an international survey on common practices related to biomedical competitions. We further linked the strategies of teams to the competition ranking in order to tackle the question “Why is the winner the best?”, which was answered more specifically in our related publication \cite{eisenmann2023winner} by addressing an additional survey solely to the competition winners.

\begin{acks}
We thank Susanne Steger (Data Protection Office, DKFZ) for the data protection supervision and Anke Trotter (IT Core Facility, DKFZ) for the hosting of the surveys.

For completing the survey, we further thank Julia Andresen (Institute for Medical Informatics, University of Lübeck, Germany), Florentin Bieder (University of Basel, Switzerland), Xue Feng (University of Virginia, USA), Alex Golts (IBM Research), Edward Henderson (The University of Manchester, UK), Yisen Huang (CUHK, Hong Kong), Ángel Víctor Juanco-Müller (Heriot-Watt University; Canon Medical Research Europe), Yunshuang Li (Zhejiang University, China), Amirreza Mahbod (Medical University of Vienna, Austria), Hans Meine (University of Bremen; Fraunhofer MEVIS, Germany), Ying Peng (Shaanxi Normal University, China), Ferran Prados Carrasco (CMIC - University College London and e-Health Center - Universitat Oberta de Catalunya, UK), Wolfgang Reiter (Wintegral GmbH, Germany), Raviteja Sista (Centre of Excellence in Artificial Intelligence, Indian Institute of Technology Kharagpur, India), Sebastian Starke (Helmholtz-Zentrum Dresden-Rossendorf, Germany), Jeremy Tan (Imperial College London, UK), Sandrine Voros (GMCAO/CAMI team at TIMC Lab UMR 5525 CNRS, UGA, Grenoble, France), Yangyang Wang (University of Missouri-Columbia, USA), Amine Yamlahi (German Cancer Research Center (DKFZ) Heidelberg, Division of Intelligent Medical Systems, Germany, Germany), and all participants who do not wish to be named.

For establishing the contact to challenge participants, we thank Marc Aubreville (Technische Hochschule Ingolstadt, Germany).

\end{acks}

\bibliographystyle{ACM-Reference-Format}
\bibliography{sample-base}

\newpage

\appendix


\section{Author list}
\label{app:authors}

The challenge organizers and chairs who contributed to our study qualified for co-authorship. Challenge participants qualified if they completed the survey and opted in for co-authorship. Respondents who participated in both an ISBI or a MICCAI challenge are listed once.

\subsection{Organizers and chairs of IEEE ISBI 2021 challenges}
\label{app:authors_organizers_isbi}
\setlength{\parindent}{0pt}

\textbf{Sharib Ali}, School of Computing, University of Leeds, Leeds, UK

\textbf{Anubha Gupta}, SBILab, Department of ECE, IIIT-Delhi, India

\textbf{Jan Kybic}, Faculty of Electrical Engineering, Czech Technical University, Prague, Czech Republic

\textbf{Alison Noble}, Institute of Biomedical Engineering, University of Oxford, UK

\textbf{Carlos Ortiz de Solórzano}, Center for Applied Medical Research, Pamplona, Spain

\textbf{Samiksha Pachade}, Shri Guru Gobind Singhji Institute of Engineering and Technology, Nanded, Maharashtra, India

\textbf{Caroline Petitjean}, Université de Rouen Normandie, France

\textbf{Daniel Sage}, School of Engineering, Biomedical Imaging Group, École Polytechnique Fédérale de Lausanne (EPFL), Lausanne, Switzerland

\textbf{Donglai Wei}, School of Engineering and Applied Science, Harvard University, USA

\textbf{Elizabeth Wilden}, Institute of Biomedical Engineering, University of Oxford, UK

\subsection{Organizers and chairs of MICCAI 2021 challenges}
\label{app:authors_organizers_miccai}
\setlength{\parindent}{0pt}

\textbf{Deepak Alapatt}, ICube, University of Strasbourg, CNRS, France

\textbf{Vincent Andrearczyk}, Institute of Informatics, School of Management, HES-SO Valais-Wallis University of Applied Sciences and Arts Western Switzerland, Techno-Pôle 3, 3960 Sierre, Switzerland; Department of Nuclear Medicine and Molecular Imaging, Lausanne University Hospital, Rue du Bugnon 46, CH-1011, Lausanne, Switzerland

\textbf{Ujjwal Baid}, Center for Biomedical Image Computing and Analytics (CBICA), University of Pennsylvania, Philadelphia, PA, USA

\textbf{Spyridon Bakas}, Center for Biomedical Image Computing and Analytics (CBICA), University of Pennsylvania, Philadelphia, PA, USA

\textbf{Niranjan Balu}, Department of Radiology, University of Washington, USA

\textbf{Sophia Bano}, Wellcome/EPSRC Centre for Interventional and Surgical Sciences, University College London, London, UK

\textbf{Vivek Singh Bawa}, Visual Artificial Intelligence Lab, Oxford Brookes University, Oxford, UK

\textbf{Jorge Bernal}, Universitat Autònoma de Barcelona \& Computer Vision Center, Spain

\textbf{Sebastian Bodenstedt}, Translational Surgical Oncology - National Center for Tumor Diseases (NCT), Partner Site Dresden, Fetscherstraße 74, PF 64, 01307 Dresden, Germany

\textbf{Alessandro Casella}, Istituto Italiano di Tecnologia, Genoa/Politecnico di Milano, Milan, Italy

\textbf{Jinwook Choi}, Seoul National University Hospital, Seoul, South Korea

\textbf{Olivier Commowick}, INRIA Rennes - Bretagne Atlantique, Empenn research team, Campus de Beaulieu, 35000 Rennes, France

\textbf{Marie Daum}, Department of General, Visceral and Transplantation Surgery, Heidelberg University Hospital, Germany

\textbf{Adrien Depeursinge}, Institute of Informatics, School of Management, HES-SO Valais-Wallis University of Applied Sciences and Arts Western Switzerland, Techno-Pôle 3, 3960 Sierre, Switzerland; Department of Nuclear Medicine and Molecular Imaging, Lausanne University Hospital, Rue du Bugnon 46, CH-1011, Lausanne, Switzerland

\textbf{Reuben Dorent}, Department of Neurosurgery, Brigham and Women’s Hospital, Harvard Medical School, Boston, MA, USA

\textbf{Jan Egger}, Institute for Artificial Intelligence in Medicine (IKIM), University Hospital Essen (AöR), Essen, Germany

\textbf{Hannah Eichhorn}, Neurobiology Research Unit, Copenhagen University Hospital, Rigshospitalet, Copenhagen, Denmark

\textbf{Sandy Engelhardt}, Department of Internal Medicine III, Heidelberg University Hospital, Heidelberg, Germany

\textbf{Melanie Ganz}, Neurobiology Research Unit, Copenhagen University Hospital, Rigshospitalet, Copenhagen; Department of Computer Science, University of Copenhagen, Copenhagen, Denmark

\textbf{Gabriel Girard}, CIBM Center for Biomedical Imaging \& Centre Hospitalier Universitaire Vaudois~\& École Polytechnique Fédérale de Lausanne, Switzerland

\textbf{Lasse Hansen}, University of Lübeck, Germany

\textbf{Mattias Heinrich}, University of Lübeck, Germany

\textbf{Nicholas Heller}, Department of Computer Science \& Engineering, University of Minnesota – Twin Cities, USA

\textbf{Alessa Hering}, Diagnostic Image Analysis Group, Radboud University Medical Center, Nijmegen, The Netherlands; Fraunhofer MEVIS, Lübeck, Germany

\textbf{Arnaud Huaulmé}, Univ Rennes, INSERM, LTSI - UMR 1099, F35000, Rennes, France

\textbf{Hyunjeong Kim}, Seoul National University Hospital, Seoul, South Korea

\textbf{Hongwei Bran Li}, University of Zurich, Switzerland; Technical University of Munich, Germany

\textbf{Bennett Landman}, Electrical Engineering, Vanderbilt University, Nashville, Tennessee, USA

\textbf{Jianning Li}, Institute for Artificial Intelligence in Medicine (IKIM), University Hospital Essen (AöR), Essen, Germany

\textbf{Jun Ma}, Nanjing University of Science and Technology, China

\textbf{Anne Martel}, Department of Medical Biophysics, University of Toronto, Ontario, Canada; Physical Sciences, Sunnybrook Research Institute, Toronto, Ontario, Canada

\textbf{Carlos Martín-Isla}, Universitat de Barcelona, Spain

\textbf{Bjoern Menze}, University of Zurich, Switzerland; Technical University of Munich, Germany

\textbf{Chinedu Innocent Nwoye}, ICube, University of Strasbourg, CNRS, France

\textbf{Valentin Oreiller}, Institute of Informatics, School of Management, HES-SO Valais-Wallis University of Applied Sciences and Arts Western Switzerland, Techno-Pôle 3, 3960 Sierre, Switzerland; Department of Nuclear Medicine and Molecular Imaging, Lausanne University Hospital, Rue du Bugnon 46, CH-1011, Lausanne, Switzerland

\textbf{Nicolas Padoy}, ICube, University of Strasbourg, CNRS, France; IHU Strasbourg, France

\textbf{Sarthak Pati}, Center For Artificial Intelligence And Data Science For Integrated Diagnostics (AI\textsuperscript{2}D) and Center for Biomedical Image Computing and Analytics (CBICA), University of Pennsylvania, Philadelphia, PA, USA; Department of Pathology and Laboratory Medicine, Perelman School of Medicine, University of Pennsylvania, Philadelphia, PA, USA; Department of Radiology, Perelman School of Medicine, University of Pennsylvania, Philadelphia, PA, USA; Department of Informatics, Technical University of Munich, Munich, Germany

\textbf{Kelly Payette}, Center for MR Research, University Children's Hospital Zurich, University of Zurich, Zurich, Switzerland

\textbf{Carole Sudre}, MRC Unit for Lifelong Health and Ageing, University College London, UK; Centre for Medical Image Computing, University College London, UK; School of Biomedical Engineering \& Imaging Sciences, King's College London, UK; Dementia Research Centre, University College London, UK

\textbf{Kimberlin van Wijnen}, Biomedical Imaging Group, Erasmus MC, The Netherlands

\textbf{Armine Vardazaryan}, IHU Strasbourg, France

\textbf{Tom Vercauteren}, King’s College London, UK

\textbf{Martin Wagner}, Department of General, Visceral and Transplantation Surgery, Heidelberg University Hospital, Germany

\textbf{Chuanbo Wang}, Big Data Analytics and Visualization Laboratory, University of Wisconsin at Milwaukee, Milwaukee, 3200 N. Cramer St, Milwaukee, WI 53211, USA

\textbf{Moi Hoon Yap}, Department of Computing and Mathematics, Manchester Metropolitan University, UK

\textbf{Zeyun Yu}, Big Data Analytics and Visualization Laboratory, University of Wisconsin at Milwaukee, Milwaukee, 3200 N. Cramer St, Milwaukee, WI 53211, USA

\textbf{Chun Yuan}, Department of Radiology, University of Washington, USA

\textbf{Maximilian Zenk}, German Cancer Research Center (DKFZ) Heidelberg, Division of Medical Image Computing, Germany

\textbf{Aneeq Zia}, Intuitive Surgical Inc, USA

\textbf{David Zimmerer}, German Cancer Research Center (DKFZ) Heidelberg, Division of Medical Image Computing, Germany

\subsection{Participants of IEEE ISBI 2021 challenges}
\label{app:authors_participants_isbi}
\setlength{\parindent}{0pt}

\textbf{Rina Bao}, University of Missouri-Columbia, USA

\textbf{Chanyeol Choi}, Massachusetts Institute of Technology, Cambridge, MA, USA

\textbf{Andrew Cohen}, Drexel University, USA

\textbf{Oleh Dzyubachyk}, N/A, The Netherlands

\textbf{Adrian Galdran}, Universidad Pompeu Fabra, Barcelona, Spain

\textbf{Tianyuan Gan}, College of Biomedical Engineering and Instrument Science, Zhejiang University, China

\textbf{Tianqi Guo}, Purdue University, USA

\textbf{Pradyumna Gupta}, Indian Institute of Technology (Indian School of Mines), Dhanbad, India

\textbf{Mahmood Haithami}, University of Nottingham, Malaysia Campus, Malaysia

\textbf{Edward Ho}, Schulich School of Medicine \& Dentistry, Western University, Canada

\textbf{Ikbeom Jang}, Massachusetts General Hospital, Boston, MA, Unites States; Harvard Medical School, Boston, MA, USA

\textbf{Zhili Li}, University of Science and Technology of China, China

\textbf{Zhengbo Luo}, Waseda University, Japan

\textbf{Filip Lux}, Centre for Biomedical Image Analysis, Masaryk University, Brno, Czech Republic

\textbf{Sokratis Makrogiannis}, Delaware State University, USA

\textbf{Dominik Müller}, IT-Infrastructure for Translational Medical Research, University of Augsburg, Germany

\textbf{Young-tack Oh}, Sungkyunkwan University, Suwon, Korea

\textbf{Subeen Pang}, Massachusetts Institute of Technology, Cambridge, MA, USA

\textbf{Constantin Pape}, Institute of Computer Science, Georg-August-Universität Göttingen, Germany

\textbf{Gorkem Polat}, Graduate School of Informatics, Middle East Technical University, Turkey

\textbf{Charlotte Rosalie Reed}, Northwestern University Feinberg School of Medicine, Chicago, IL, USA

\textbf{Kanghyun Ryu}, Stanford University, Palo Alto, CA, USA

\textbf{Tim Scherr}, Institute for Automation and Applied Informatics, Karlsruhe Institute of Technology, Eggenstein-Leopoldshafen, Germany

\textbf{Vajira Thambawita}, SimulaMet, Norway

\textbf{Haoyu Wang}, Shanghai Jiao Tong University, China

\textbf{Xinliang Wang}, Beihang University, China

\textbf{Kele Xu}, National University of Defense Technology, China

\textbf{Hung Yeh}, National United University, Taiwan

\textbf{Doyeob Yeo}, Korea Atomic Energy Research Institute, South Korea

\textbf{Yixuan Yuan}, City University of Hong Kong, Hong Kong

\textbf{Yan Zeng}, Beijing University of Technology, Beijing, China

\textbf{Xin Zhao}, The Center for Research on Intelligent System and Engineering, Institute of Automation, Chinese Academy of Sciences, China

\subsection{Participants of MICCAI 2021 challenges}
\label{app:authors_participants_miccai}
\setlength{\parindent}{0pt}


\textbf{Julian Abbing}, University of Twente; Johnson and Johnson, The Netherlands

\textbf{Jannes Adam}, University of Bremen, Germany

\textbf{Nagesh Adluru}, University of Wisconsin, Madison, USA

\textbf{Niklas Agethen}, University of Bremen, Germany

\textbf{Salman Ahmed}, National University of Computer and Emerging Sciences, Islamabad, Pakistan

\textbf{Yasmina Al Khalil}, Eindhoven University of Technology, Eindhoven, The Netherlands

\textbf{Deepak Alapatt}, ICube, University of Strasbourg, France

\textbf{Mireia Alenyà}, BCN-MedTech, Department of Information and Communications Technologies, Universitat Pompeu Fabra, Spain

\textbf{Esa Alhoniemi}, Turku University of Applied Sciences, Finland

\textbf{Chengyang An}, Shanghai Jiao Tong University, Shanghai, China

\textbf{Talha Anwar}, N/A

\textbf{Tewodros Weldebirhan Arega}, ImViA Laboratory, Université Bourgogne Franche-Comté, Dijon, France

\textbf{Netanell Avisdris}, School of Computer Science and Engineering, Hebrew University of Jerusalem, Israel; Sagol Brain Institute, Tel Aviv Sourasky Medical Center, Israel

\textbf{Dogu Baran Aydogan}, A.I. Virtanen Institute for Molecular Sciences, University of Eastern Finland, Finland

\textbf{Yingbin Bai}, The University of Sydney, Australia

\textbf{Maria Baldeon Calisto}, Universidad San Francisco de Quito, Ecuador

\textbf{Berke Doga Basaran}, Department of Computing, Imperial College London, UK

\textbf{Marcel Beetz}, University of Oxford, UK

\textbf{Cheng Bian}, N/A

\textbf{Hao Bian}, Tsinghua University, China

\textbf{Kevin Blansit}, Subtle Medical, USA

\textbf{Louise Bloch}, Department of Computer Science, University of Applied Sciences and Arts Dortmund, Dortmund; Institute for Medical Informatics, Biometry and Epidemiology (IMIBE), University Hospital Essen, Essen; Institute for Artificial Intelligence in Medicine (IKIM), University Hospital Essen, Essen, Germany

\textbf{Robert Bohnsack}, University of Bremen, Germany

\textbf{Sara Bosticardo}, University of Verona, Italy

\textbf{Jack Breen}, University of Leeds, UK

\textbf{Mikael Brudfors}, Wellcome Centre for Human Neuroimaging, University College London, London, UK

\textbf{Raphael Brüngel}, Department of Computer Science, University of Applied Sciences and Arts Dortmund, Dortmund; Institute for Medical Informatics, Biometry and Epidemiology (IMIBE), University Hospital Essen, Essen; Institute for Artificial Intelligence in Medicine (IKIM), University Hospital Essen, Essen, Germany

\textbf{Mariano Cabezas}, Brain and Mind Centre, The University of Sydney, Australia

\textbf{Alberto Cacciola}, Brain Mapping Lab, Department of Biomedical, Dental Sciences and Morphological and Functional Images, University of Messina, Messina, Italy

\textbf{Daniel Tianming Chen}, unaffiliated, Canada

\textbf{Yucong Chen}, ShanghaiTech University, China

\textbf{Zhiwei Chen}, Shaanxi Normal University, China

\textbf{Minjeong Cho}, Chung-Ang University, South Korea

\textbf{Min-Kook Choi}, VisionAI, hutom, South Korea

\textbf{Dana Cobzas}, MacEwan University, Canada; University of Alberta, Canada

\textbf{Julien Cohen-Adad}, NeuroPoly Lab, Institute of Biomedical Engineering, Polytechnique Montréal, Montréal, QC, Canada; MILA - Québec AI Institute, Montréal, QC, Canada

\textbf{Jorge Corral Acero}, University of Oxford, UK

\textbf{Sujit Kumar Das}, Department of Computer Science and Engineering, National Institute of Technology Silchar, India

\textbf{Marcela de Oliveira}, São Paulo State University (UNESP), Brazil

\textbf{Hanqiu Deng}, Department of Electrical and Computer Engineering, University of Alberta, Canada

\textbf{Guiming Dong}, University of Electronic Science and Technology of China, China

\textbf{Lars Doorenbos}, University of Bern, Switzerland

\textbf{Cory Efird}, MacEwan University, Canada; University of Alberta, Canada

\textbf{Sergio Escalera}, University of Barcelona, Spain

\textbf{Di Fan}, University of California, Irvine, USA

\textbf{Mehdi Fatan Serj}, Institut de Robòtica i Informàtica Industrial, Spain

\textbf{Alexandre Fenneteau}, XLIM Laboratory, University of Poitiers, UMR CNRS 7252 - Poitiers, France; I3M, Common Laboratory CNRS-Siemens - University and Hospital of Poitiers ; Poitiers, France - Neuroimaging Department, Quinze Vingts Hospital; Paris, France; Siemens Healthcare, Saint Denis, France

\textbf{Lucas Fidon}, School of Biomedical Engineering \& Imaging Sciences, King’s College London, UK

\textbf{Patryk Filipiak}, Center for Advanced Imaging Innovation and Research (CAI2R), Department of Radiology, NYU Langone Health, New York, NY, USA

\textbf{René Finzel}, University of Bremen, Germany

\textbf{Nuno R. Freitas}, CMEMS-Uminho, University of Minho, Portugal

\textbf{Christoph M. Friedrich}, Department of Computer Science, University of Applied Sciences and Arts Dortmund, Dortmund; Institute for Medical Informatics, Biometry and Epidemiology (IMIBE), University Hospital Essen, Essen, Germany

\textbf{Mitchell Fulton}, University of Colorado, Boulder, USA

\textbf{Finn Gaida}, Technical University Munich, Germany

\textbf{Francesco Galati}, EURECOM, France

\textbf{Christoforos Galazis}, Department of Computing, Imperial College London and National Heart and Lung Institute, Imperial College London, UK

\textbf{Adrian Galdran}, Bournemouth University, UK

\textbf{Chang Hee Gan}, Sejong University/Korea University, South Korea

\textbf{Shengbo Gao}, Beihang University, China

\textbf{Zheyao Gao}, N/A

\textbf{Matej Gazda}, Technical University of Kosice, Slovakia

\textbf{Beerend Gerats}, Centre for Artificial Intelligence, Meander Medisch Centrum, The Netherlands

\textbf{Neil Getty}, Government Research, USA

\textbf{Adam Gibicar}, Ryerson University, Canada

\textbf{Ryan Gifford}, Ohio State University, USA

\textbf{Sajan Gohil}, Pandit Deendayal Energy University, India

\textbf{Maria Grammatikopoulou}, Digital Surgery, Medtronic, UK

\textbf{Daniel Grzech}, Imperial College London, UK

\textbf{Chunxu Guo}, ShanghaiTech University School of Biomedical Engineering IDEALab, China

\textbf{Sylvain Guy}, GMCAO/CAMI team at TIMC Lab UMR 5525 CNRS, UGA, Grenoble, France

\textbf{Orhun Güley}, N/A

\textbf{Timo Günnemann}, University of Bremen, Germany

\textbf{Heonjin Ha}, Saige Research, South Korea

\textbf{Il Song Han}, N/A

\textbf{Luyi Han}, NKI, The Netherlands

\textbf{Ali Hatamizadeh}, NVIDIA, N/A

\textbf{Tian He}, Fuzhou University, China

\textbf{Jimin Heo}, N/A

\textbf{Sebastian Hitziger}, Mediaire, Germany

\textbf{SeulGi Hong}, VisionAI, hutom, South Korea

\textbf{SeungBum Hong}, VisionAI, hutom, South Korea

\textbf{Rian Huang}, National-Regional Key Technology Engineering Laboratory for Medical Ultrasound, Guangdong Key Laboratory for Biomedical Measurements and Ultrasound Imaging, School of Biomedical Engineering, Health Science Center, Shenzhen University, Shenzhen, China

\textbf{Ziyan Huang}, Shanghai Jiao Tong University, China

\textbf{Markus Huellebrand}, Charité - Universitätsmedizin Berlin, Germany

\textbf{Stephan Huschauer}, unaffiliated, N/A

\textbf{Mustaffa Hussain}, Onward Assist, India

\textbf{Tomoo Inubushi}, Hamamatsu Photonics K.K., Japan

\textbf{Ece Isik Polat}, Graduate School of Informatics, Middle East Technical University, Turkey

\textbf{Mojtaba Jafaritadi}, Stanford University, Department of Radiology, California, USA; Turku University of Applied Sciences, Faculty of Engineering and Business, Turku, Finland

\textbf{SeongHun Jeong}, Korea Telecom, South Korea

\textbf{Bailiang Jian}, Technical University of Munich, Germany

\textbf{Yuanhong Jiang}, Shanghai Jiao Tong University, China

\textbf{Zhifan Jiang}, Sheikh Zayed Institute for Pediatric Surgical Innovation, Children's National Hospital, Washington, DC, USA

\textbf{Yueming Jin}, University College London, UK

\textbf{Smriti Joshi}, University of Barcelona, Spain

\textbf{Abdolrahim Kadkhodamohammadi}, Digital Surgery, Medtronic, UK

\textbf{Reda Abdellah Kamraoui}, Université de Bordeaux, France

\textbf{Inha Kang}, Korea Advanced Institute of Science and Technology, South Korea

\textbf{Junghwa Kang}, Hankuk University of Foreign Studies, South Korea

\textbf{Davood Karimi}, Harvard Medical School, USA

\textbf{April Khademi}, Image Analysis in Medicine Lab (IAMLAB), Department of Electrical, Computer, and Biomedical Engineering, Toronto Metropolitan University (formerly Ryerson University) Toronto, Canada; Keenan Research Center for Biomedical Science, St. Michael's Hospital, Unity Health Network, Toronto, Canada; Institute for Biomedical Engineering, Science, and Technology (iBEST), A Partnership Between St. Michael's Hospital and TMU, Toronto, Canada

\textbf{Muhammad Irfan Khan}, Turku University of Applied Sciences, Finland

\textbf{Suleiman A. Khan}, Turku University of Applied Sciences, Finland

\textbf{Rishab Khantwal}, Centre for Machine Intelligence and Data Science, Indian Institute of Technology Bombay, India

\textbf{Kwang-Ju Kim}, Electronics and Telecommunications Research Institute, South Korea

\textbf{Timothy Kline}, Mayo Clinic, USA

\textbf{Satoshi Kondo}, Muroran Institute of Technology, Japan

\textbf{Elina Kontio}, Turku University of Applied Sciences, Finland

\textbf{Adrian Krenzer}, University of Würzburg, Germany

\textbf{Artem Kroviakov}, Institute of Computer Graphics and Vision, Graz University of Technology, Austria

\textbf{Hugo Kuijf}, UMC Utrecht, The Netherlands

\textbf{Satyadwyoom Kumar}, N/A, India

\textbf{Francesco La Rosa}, Icahn School of Medicine at Mount Sinai, USA

\textbf{Abhi Lad}, Pandit Deendayal Energy University, India

\textbf{Doohee Lee}, ZIOVISION Co., Ltd., South Korea

\textbf{Minho Lee}, Neurophet Inc., South Korea

\textbf{Chiara Lena}, Politecnico di Milano, Italy

\textbf{Hao Li}, Vanderbilt University, USA

\textbf{Jianning Li}, Institute for Artificial Intelligence in Medicine (IKIM), University Hospital Essen (AöR), Essen, Germany

\textbf{Ling Li}, School of Management, Hefei University of Technology, Hefei, China

\textbf{Xingyu Li}, University of Alberta, Canada

\textbf{Fuyuan Liao}, Xi'an Technological Unoversity, China

\textbf{KuanLun Liao}, Real Doctor AI Research Centre, China

\textbf{Arlindo Limede Oliveira}, INESC-ID / Instituto Superior Técnico, University of Lisbon, Portugal

\textbf{Chaonan Lin}, Fuzhou University, China

\textbf{Shan Lin}, University of California San Diego, USA

\textbf{Akis Linardos}, University of Barcelona, Spain

\textbf{Marius George Linguraru}, Sheikh Zayed Institute for Pediatric Surgical Innovation, Children's National Hospital, Washington, DC, USA ; School of Medicine and Health Sciences, George Washington University, Washington, DC, USA

\textbf{Di Liu}, Rutgers University, USA

\textbf{Han Liu}, Vanderbilt University, Nashville, Tennessee, USA

\textbf{Tao Liu}, Jiangnan University, China

\textbf{Yanling Liu}, Leidos Biomed., USA

\textbf{João Lourenço-Silva}, INESC-ID / Instituto Superior Técnico, University of Lisbon, Portugal

\textbf{Jiangshan Lu}, University of Electronic Science and Technology of China, China

\textbf{Jingpei Lu}, University of California San Diego, USA

\textbf{Imanol Luengo}, Digital Surgery, Medtronic, UK

\textbf{Christina B. Lund}, Image Sciences Institute, UMC Utrecht, Utrecht University, The Netherlands

\textbf{Huan Minh Luu}, MRI Lab, Department of Bio and Brain Engineering, Korea Advanced Institute of Science and Technology, N/A

\textbf{Yi Lv}, Beihang University, China

\textbf{Yi Lv}, Beijing Institute of Technology, China

\textbf{Uzay Macar}, NeuroPoly Lab, Institute of Biomedical Engineering, Polytechnique Montréal, Montréal, QC, Canada; MILA - Québec AI Institute, Montréal, QC, Canada

\textbf{Leon Maechler}, ENS Paris, France

\textbf{Sina Mansour L.}, Department of Biomedical Engineering, The University of Melbourne, Parkville, Victoria, Australia

\textbf{Kenji Marshall}, McGill University, Canada

\textbf{Moona Mazher}, Department of Computer Engineering and Mathematics, University Rovira i Virgili, Spain

\textbf{Richard McKinley}, Support Centre for Advanced Neuroimaging, University institute for diagnostic and interventional neuroradiology, university hospital Bern, Switzerland

\textbf{Alfonso Medela}, Legit Health, Spain

\textbf{Felix Meissen}, Klinikum rechts der Isar of the Technical University of Munich, Germany

\textbf{Mingyuan Meng}, The University of Sydney, Australia

\textbf{Dylan Miller}, MacEwan University, Canada; University of Alberta, Canada

\textbf{Seyed Hossein Mirjahanmardi}, Ryerson University, Canada

\textbf{Arnab Mishra}, Department of Computer Science and Engineering, National Institute of Technology Silchar, India

\textbf{Samir Mitha}, Department of Electrical, Computer and Biomedical Engineering, Ryerson University, Toronto, ON, Canada

\textbf{Hassan Mohy-ud-Din}, LUMS School of Science and Engineering, Lahore, Pakistan

\textbf{Tony Chi Wing Mok}, The Hong Kong University of Science and Technology, Hong Kong

\textbf{Gowtham Krishnan Murugesan}, BAMF Health, USA

\textbf{Enamundram Naga Karthik}, NeuroPoly Lab, Institute of Biomedical Engineering, Polytechnique Montréal, Montréal, QC, Canada; MILA - Québec AI Institute, Montréal, QC, Canada

\textbf{Sahil Nalawade}, UT Southwestern Medical Center, USA

\textbf{Jakub Nalepa}, Silesian University of Technology, Gliwice, Poland; Graylight Imaging, Gliwice, Poland

\textbf{Mohamed Naser}, MD Anderson Cancer Center, USA

\textbf{Ramin Nateghi}, Department of Electrical and Electronics Engineering, Shiraz University of Technology, Shiraz, Iran

\textbf{Hammad Naveed}, National University of Computer and Emerging Sciences, Islamabad, Pakistan

\textbf{Quang-Minh Nguyen}, Laboratoire du Traitement du Signal et de l'Image - INSERM UMR 1099, France

\textbf{Cuong Nguyen Quoc}, University of Information Technology, Vietnam National University Ho Chi Minh City, Vietnam

\textbf{Brennan Nichyporuk}, McGill University, Canada

\textbf{Bruno Oliveira}, 2Ai – School of Technology, IPCA, Barcelos, Portugal; Algoritmi Center, School of Engineering, University of Minho, Guimarães, Portugal; Life and Health Sciences Research Institute (ICVS), School of Medicine, University of Minho, Braga, Portugal; ICVS/3B’s - PT Government Associate Laboratory, Braga/Guimarães, Portugal

\textbf{David Owen}, Digital Surgery, Medtronic, UK

\textbf{Jimut Bahan Pal}, Centre for Machine Intelligence and Data Science, Indian Institute of Technology, Bombay, India

\textbf{Junwen Pan}, Tianjin University, China

\textbf{Wentao Pan}, Tsinghua University, China

\textbf{Winnie Pang}, National University of Singapore, Singapore

\textbf{Bogyu Park}, Hutom Inc., South Korea

\textbf{Kamlesh Pawar}, Monash University, Australia

\textbf{Vivek Pawar}, Endimension Tech., India

\textbf{Michael Peven}, Johns Hopkins University, USA

\textbf{Lena Philipp}, University of Bremen, Germany

\textbf{Tomasz Pieciak}, LPI, ETSI Telecomunicación, Universidad de Valladolid, Valladolid, Spain

\textbf{Szymon Plotka}, Sano Centre for Computational Medicine, Cracow, Poland

\textbf{Marcel Plutat}, University of Bremen, Germany

\textbf{Fattaneh Pourakpour}, Iranian Brain Mapping Biobank(IBMB), National Brain Mapping Laboratory (NBML), Tehran, Iran

\textbf{Domen Preložnik}, Laboratory of Imaging Technologies - Faculty of Electrical Engineering, University of Ljubljana, Slovenia

\textbf{Kumaradevan Punithakumar}, Department of Radiology and Diagnostic Imaging, University of Alberta, Canada

\textbf{Abdul Qayyum}, School of Biomedical Engineering \& Imaging Sciences, King's College London, UK

\textbf{Sandro Queirós}, Life and Health Sciences Research Institute (ICVS), School of Medicine, University of Minho, Braga, Portugal; ICVS/3B’s - PT Government Associate Laboratory, Braga/Guimarães, Portugal

\textbf{Arman Rahmim}, University of British Columbia, Canada

\textbf{Salar Razavi}, Ryerson University, Canada

\textbf{Jintao Ren}, Department of Oncology, Aarhus University, Denmark

\textbf{Mina Rezaei}, LMU Munich, Germany

\textbf{Jonathan Adam Rico}, University of San Agustin, Center for Informatics, Iloilo City, Philippines

\textbf{ZunHyan Rieu}, Research Institute, NEUROPHET Inc., Seoul 06247, South Korea

\textbf{Markus Rink}, University of Bremen, Germany

\textbf{Johannes Roth}, Leipzig University, Germany

\textbf{Yusely Ruiz-Gonzalez}, Universidad Central "Marta Abreu" de Las Villas, Cuba

\textbf{Numan Saeed}, Mohamed Bin Zayed University of Science and Technology, United Arab Emirates

\textbf{Anindo Saha}, Diagnostic Image Analysis Group, Radboud University Medical Center, The Netherlands

\textbf{Mostafa Salem}, Institute of Computer Vision and Robotics, University of Girona, Girona, Spain; Computer Science Department, Faculty of Computers and Information, Assiut University, Egypt

\textbf{Ricardo Sanchez-Matilla}, Digital Surgery, Medtronic, UK

\textbf{Kurt Schilling}, Vanderbilt University Medical Center, USA

\textbf{Wei Shao}, Department of Radiology, Stanford University, USA

\textbf{Zhiqiang Shen}, Fuzhou University, China

\textbf{Pengcheng Shi}, Harbin Institute of Technology, Shenzhen, China

\textbf{Ruize Shi}, N/A, China

\textbf{Daniel Sobotka}, Computational Imaging Research Lab, Department of Biomedical Imaging and Image-guided Therapy, Medical University of Vienna, Vienna, Austria

\textbf{Théodore Soulier}, Paris Brain Institute, France

\textbf{Bella Specktor Fadida}, School of Computer Science and Engineering, Hebrew University of Jerusalem, Israel

\textbf{Danail Stoyanov}, Digital Surgery, Medtronic; University College London, UK

\textbf{Timothy Sum Hon Mun}, Institute of Cancer Research, London, UK

\textbf{Xiaowu Sun}, Leiden University Medical Center, The Netherlands

\textbf{Rong Tao}, Institute of Medical Robotics, Shanghai Jiao Tong University, China

\textbf{Franz Thaler}, Institute of Computer Graphics and Vision, Graz University of Technology; Division of Medical Physics and Biophysics, Medical University of Graz, Austria

\textbf{Felix Thielke}, University of Bremen; Fraunhofer MEVIS, Germany

\textbf{Antoine Théberge}, Université de Sherbrooke, Canada

\textbf{Helena Torres}, 2Ai – School of Technology, IPCA, Barcelos, Portugal; Algoritmi Center, School of Engineering, University of Minho, Guimarães, Portugal; Life and Health Sciences Research Institute (ICVS), School of Medicine, University of Minho, Braga, Portugal; ICVS/3B’s - PT Government Associate, Portugal

\textbf{Kareem A. Wahid}, MD Anderson Cancer Center, USA

\textbf{Jiacheng Wang}, Xiamen University, China

\textbf{Wei Wang}, Southern University of Science and Technology, China

\textbf{Xiong Wang}, OPPO Research Institute, China

\textbf{YiFei Wang}, N/A

\textbf{Jianhui Wen}, N/A

\textbf{Ning Wen}, Henry Ford Health System, USA

\textbf{Marek Wodzinski}, AGH UST, Department of Measurement and Electronics, Poland; University of Applied Sciences and Arts Western Switzerland (HES-SO), Switzerland

\textbf{Ye Wu}, Nanjing University of Science and Technology, China

\textbf{Tianqi Xiang}, The Hong Kong University of Science and Technology, Hong Kong

\textbf{Fangfang Xia}, University of Chicago, USA

\textbf{Chen Xiaofei}, Southeast University, India

\textbf{Chuantao Xie}, Shanghai University, China

\textbf{Lizhan Xu}, N/A, China

\textbf{Tingting Xue}, University of Bremen, Germany

\textbf{Lin Yang}, The Hong Kong University of Science and Technology, Hong Kong

\textbf{Yuxuan Yang}, School of Management, Hefei University of Technology, China

\textbf{Huifeng Yao}, Shandong University, China

\textbf{Kai Yao}, School of Advanced Technology, Xi’an Jiaotong-Liverpool University, China

\textbf{Amirsaeed Yazdani}, The Pennsylvania State University, USA

\textbf{Michael Yip}, University of California San Diego, USA

\textbf{Hwanseung Yoo}, Hankuk University of Foreign Studies, South Korea

\textbf{Fereshteh Yousefirizi}, BC Cancer Research Institute, Canada

\textbf{Lei Yu}, Graz University of technology, Austria

\textbf{Shunkai Yu}, University of California San Diego, USA

\textbf{Qingshuo Zheng}, Surgical Robot Vision Group, Wellcome/EPSRC Centre for Interventional and Surgical Sciences(WEISS) and Department of Computer Science, University College London, London, UK

\textbf{Jonathan Zamora}, University of California San Diego, USA

\textbf{Ramy Ashraf Zeineldin}, Health Robotics and Automation (HERA), Karlsruhe Institute of Technology, Germany; Research Group Computer Assisted Medicine (CaMed), Reutlingen University, Germany; Faculty of Electronic Engineering (FEE), Menoufia University, Egypt

\textbf{Dewen Zeng}, University of Notre Dame, USA

\textbf{Bokai Zhang}, Johnson \& Johnson, N/A

\textbf{Fan Zhang}, Fosun Aitrox Information Technology Ltd., China

\textbf{Huahong Zhang}, Vanderbilt University, USA

\textbf{Jianpeng Zhang}, Northwestern Polytechnical University, China

\textbf{Jiapeng Zhang}, University of Shanghai for Science and Technology, China

\textbf{Can Zhao}, NVIDIA, Bethesda, MD, USA

\textbf{Jiachen Zhao}, The Hong Kong University of Science and Technology, Hong Kong

\textbf{Zhongchen Zhao}, Shanghai Jiao Tong University, China

\textbf{Zixuan Zhao}, University of Chicago, USA

\textbf{Yuheng Zhi}, University of California San Diego, USA

\textbf{Ziqi Zhou}, Shenzhen University, China

\textbf{Baosheng Zou}, N/A

\newpage

\section{Overview of conferences, challenges, and competitions}
\label{app:overview}
\begin{longtable}{ p{0.5cm} p{0.8cm} p{2.3cm} p{2cm} p{5cm} }
\label{tab:overview:conferences} \\
\caption{Overview of conferences included in this meta-study.} \\
\toprule
\# & ID & Conference & Date & Conference full name \\* \midrule
\endfirsthead
\multicolumn{5}{c}%
{{\bfseries Table \thetable\ continued from previous page}} \\
\toprule
\# & ID & Conference & Date & Conference full name \\* \midrule
\endhead
1 & I & IEEE ISBI 2021 & 2021-04-13 to 2021-04-16 & 18th International Symposium on Biomedical Imaging \\
2 & M & MICCAI 2021 & 2021-09-27 to 2021-10-01 & 24th International Conference on Medical Image Computing and Computer Assisted Intervention \\* \bottomrule
\end{longtable}

\begin{longtable}{ p{0.5cm} p{0.8cm} p{2.3cm} p{2.5cm} p{6cm} }
\label{tab:overview:challenges} \\
\caption{Overview of challenges included in this meta-study.} \\
\toprule
\# & ID & Conference & Challenge acronym & Challenge full name \\* \midrule
\endfirsthead
\multicolumn{5}{c}%
{{\bfseries Table \thetable\ continued from previous page}} \\
\toprule
\# & ID & Conference & Challenge acronym & Challenge full name \\* \midrule
\endhead
\bottomrule
\endfoot
\endlastfoot
1 & I.1 & IEEE ISBI 2021 & CTC & 6th ISBI Cell Tracking Challenge \\
2 & I.2 & IEEE ISBI 2021 & MitoEM & Large-scale 3D Mitochondria Instance Segmentation Challenge \\
3 & I.3 & IEEE ISBI 2021 & EndoCV2021 & Addressing generalisability in polyp detection and segmentation challenge \\
4 & I.4 & IEEE ISBI 2021 & RIADD & Retinal Image Analysis for multi-Disease Detection Challenge \\
5 & I.5 & IEEE ISBI 2021 & SegPC-2021 & Segmentation of Multiple Myeloma Plasma Cells in Microscopic Images   Challenge \\
6 & I.6 & IEEE ISBI 2021 & A-AFMA & Ultrasound Challenge: Automatic amniotic fluid measurement and analysis from ultrasound video \\
7 & M.1 & MICCAI 2021 & KiTS21 & 2021 Kidney and Kidney Tumor Segmentation \\
8 & M.2 & MICCAI 2021 & RealNoiseMRI & Brain MRI reconstruction challenge with realistic noise \\
9 & M.3 & MICCAI 2021 & crossMoDA & Cross-Modality Domain Adaptation for Medical Image Segmentation \\
10 & M.4 & MICCAI 2021 & AdaptOR 2021 & Deep Generative Model Challenge for Domain Adaptation in Surgery 2021 \\
11 & M.5 & MICCAI 2021 & DFUC 2021 & Diabetic Foot Ulcer Challenge 2021 \\
12 & M.6a & MICCAI 2021 & HeiSurf & Endoscopic Vision Challenge 2021 - HeiChole Surgical Workflow Analysis and Full Scene Segmentation \\
13 & M.6b & MICCAI 2021 & GIANA & Endoscopic Vision Challenge 2021 - Gastrointestinal Image ANAlysis \\
14 & M.6c & MICCAI 2021 & CholecTriplet2021 & Endoscopic Vision Challenge 2021 - Surgical Action Triplet Recognition \\
15 & M.6d & MICCAI 2021 & FetReg & Endoscopic Vision Challenge 2021 - Placental Vessel Segmentation and Registration in Fetoscopy \\
16 & M.6e & MICCAI 2021 & PETRAW & Endoscopic Vision Challenge 2021 - PEg TRAnsfer Workflow recognition by different modalities \\
17 & M.6f & MICCAI 2021 & SimSurgSkill & Endoscopic Vision Challenge 2021 - Objective Surgical Skill Assessment in VR Simulation \\
18 & M.7 & MICCAI 2021 & DiSCo & Diffusion-Simulated Connectivity Challenge \\
19 & M.8 & MICCAI 2021 & FLARE21 & Fast and Low GPU Memory Abdominal Organ Segmentation in CT \\
20 & M.9 & MICCAI 2021 & FeTS & Federated Tumor Segmentation Challenge \\
21 & M.10 & MICCAI 2021 & FeTA & Fetal Brain Tissue Annotation and Segmentation Challenge \\
22 & M.11 & MICCAI 2021 & HECKTOR & HEad and neCK TumOR segmentation and outcome prediction in PET/CT images \\
23 & M.12 & MICCAI 2021 & LEARN2REG & Learn2Reg - The Challenge (2021) \\
24 & M.13 & MICCAI 2021 & MOOD & Medical Out-of-Distribution Analysis Challenge 2021 \\
25 & M.14 & MICCAI 2021 & MIDOG & MItosis DOmain Generalization Challenge 2021 \\
26 & M.15 & MICCAI 2021 & M\&Ms-2 & Multi-Disease, Multi-View \& Multi-Center Right Ventricular   Segmentation in Cardiac MRI \\
27 & M.16 & MICCAI 2021 & QUBIQ 2021 & Quantification of Uncertainties in Biomedical Image Quantification 2021 \\
28 & M.17 & MICCAI 2021 & BraTS2021 & RSNA/ASNR/MICCAI Brain Tumor Segmentation Challenge 2021 \\
29 & M.18 & MICCAI 2021 & SARAS-MESAD & SARAS challenge for Multi-domain Endoscopic Surgeon Action Detection \\
30 & M.19 & MICCAI 2021 & AutoImplant 2021 & Towards the Automatization of Cranial Implant Design in Cranioplasty: 2nd   MICCAI Challenge on Automatic Cranial Implant Design \\
31 & M.20 & MICCAI 2021 & VALDO & VAscular Lesions DetectiOn Challenge \\
32 & M.21 & MICCAI 2021 & VWS & Carotid Artery Vessel Wall Segmentation Challenge \\
33 & M.22 & MICCAI 2021 & FU-Seg & Foot Ulcer Segmentation Challenge 2021 \\
34 & M.23 & MICCAI 2021 & MSSEG-2 & Multiple sclerosis new lesions segmentation challenge \\
35 & M.24 & MICCAI 2021 & PAIP2021 & Perineural Invasion in Multiple Organ Cancer (Colon, Prostate, and   Pancreatobiliary tract) \\* \bottomrule
\end{longtable}

\begin{longtable}{ p{0.5cm} p{0.8cm} p{2.3cm} p{2.5cm} p{6cm} }
\label{tab:overview:competitions} \\
\caption{Overview of competitions included in this meta-study.} \\
\toprule
\# & ID & Conference & Challenge & Competition \\* \midrule
\endfirsthead
\multicolumn{5}{c}%
{{\bfseries Table \thetable\ continued from previous page}} \\
\toprule
\# & ID & Conference & Challenge & Competition \\* \midrule
\endhead
\bottomrule
\endfoot
\endlastfoot
1 & I.1.1 & IEEE ISBI 2021 & CTC & Primary Track (evaluation across all 13 datasets) \\
2 & I.1.2 & IEEE ISBI 2021 & CTC & Secondary Track - Dataset "DIC-C2DH-HeLa" \\
3 & I.1.3 & IEEE ISBI 2021 & CTC & Secondary Track - Dataset "Fluo-C2DL-MSC" \\
4 & I.1.4 & IEEE ISBI 2021 & CTC & Secondary Track - Dataset "Fluo-C3DH-H157" \\
5 & I.1.5 & IEEE ISBI 2021 & CTC & Secondary Track - Dataset "Fluo-C3DL-MDA231" \\
6 & I.1.6 & IEEE ISBI 2021 & CTC & Secondary Track - Dataset "Fluo-N2DH-GOWT1" \\
7 & I.1.7 & IEEE ISBI 2021 & CTC & Secondary Track - Dataset "Fluo-N2DL-HeLa" \\
8 & I.1.8 & IEEE ISBI 2021 & CTC & Secondary Track - Dataset "Fluo-N3DH-CE" \\
9 & I.1.9 & IEEE ISBI 2021 & CTC & Secondary Track - Dataset "Fluo-N3DH-CHO" \\
10 & I.1.10 & IEEE ISBI 2021 & CTC & Secondary Track - Dataset "Fluo-N3DL-DRO" \\
11 & I.1.11 & IEEE ISBI 2021 & CTC & Secondary Track - Dataset "PhC-C2DH-U373" \\
12 & I.1.12 & IEEE ISBI 2021 & CTC & Secondary Track - Dataset "PhC-C2DL-PSC" \\
13 & I.1.13 & IEEE ISBI 2021 & CTC & Secondary Track - Dataset "Fluo-N2DH-SIM+" \\
14 & I.1.14 & IEEE ISBI 2021 & CTC & Secondary Track - Dataset "Fluo-N3DH-SIM+" \\
15 & I.1.15 & IEEE ISBI 2021 & CTC & Secondary Track - Dataset "BF-C2DL-HSC" \\
16 & I.1.16 & IEEE ISBI 2021 & CTC & Secondary Track - Dataset "BF-C2DL-MuSC" \\
17 & I.1.17 & IEEE ISBI 2021 & CTC & Secondary Track - Dataset "Fluo-C2DL-Huh7" \\
18 & I.1.18 & IEEE ISBI 2021 & CTC & Secondary Track - Dataset "Fluo-C3DH-A549" \\
19 & I.1.19 & IEEE ISBI 2021 & CTC & Secondary Track - Dataset "Fluo-N3DL-TRIC" \\
20 & I.1.20 & IEEE ISBI 2021 & CTC & Secondary Track - Dataset "Fluo-N3DL-TRIF" \\
21 & I.1.21 & IEEE ISBI 2021 & CTC & Secondary Track - Dataset "Fluo-C3Dh-A549-SIM" \\
22 & I.2.1 & IEEE ISBI 2021 & MitoEM & 3D Mitochondria Instance Segmentation \\
23 & I.3.1 & IEEE ISBI 2021 & EndoCV2021 & Assessing generalisability in polyp detection \\
24 & I.3.2 & IEEE ISBI 2021 & EndoCV2021 & Assessing generalisability in polyp segmentation \\
25 & I.4.1 & IEEE ISBI 2021 & RIADD & Disease Screening \\
26 & I.4.2 & IEEE ISBI 2021 & RIADD & Disease Classification \\
27 & I.5.1 & IEEE ISBI 2021 & SegPC-2021 & Segmentation of Multiple Myeloma Plasma Cells in Microscopic Images   Challenge \\
28 & I.6.1 & IEEE ISBI 2021 & A-AFMA & Detection: Automatic amniotic fluid detection from ultrasound video \\
29 & I.6.2 & IEEE ISBI 2021 & A-AFMA & Localization: Automatic amniotic fluid measurement from ultrasound video \\
30 & M.1.1 & MICCAI 2021 & KiTS21 & Segmentation of Kidney and Associated Structures \\
31 & M.2.1 & MICCAI 2021 & RealNoiseMRI & Reconstruction of motion corrupted T1 weighted MRI data \\
32 & M.2.2 & MICCAI 2021 & RealNoiseMRI & Reconstruction of motion corrupted T2 weighted MRI data \\
33 & M.3.1 & MICCAI 2021 & crossMoDA & Vestibular Schwannoma and Cochlea Segmentation \\
34 & M.4.1 & MICCAI 2021 & AdaptOR 2021 & Domain Adaptation for Landmark Detection \\
35 & M.5.1 & MICCAI 2021 & DFUC 2021 & Analysis Towards Classification of Infection \& Ischaemia of Diabetic   Foot Ulcers \\
36 & M.6a.1 & MICCAI 2021 & HeiSurf & Scene segmentation \\
37 & M.6a.2 & MICCAI 2021 & HeiSurf & Phase segmentation \\
38 & M.6a.3 & MICCAI 2021 & HeiSurf & Instrument presence \\
39 & M.6a.4 & MICCAI 2021 & HeiSurf & Action recognition \\
40 & M.6b.1 & MICCAI 2021 & GIANA & Polyp detection in colonoscopy images \\
41 & M.6b.2 & MICCAI 2021 & GIANA & Polyp segmentation in colonoscopy images \\
42 & M.6b.3 & MICCAI 2021 & GIANA & Histology prediction \\
43 & M.6c.1 & MICCAI 2021 & CholecTriplet2021 & Surgical Action Triplet Recognition \\
44 & M.6d.1 & MICCAI 2021 & FetReg & Placental semantic segmentation \\
45 & M.6d.2 & MICCAI 2021 & FetReg & Placental RGB frame registration for mosaicking \\
46 & M.6e.1 & MICCAI 2021 & PETRAW & Video-based surgical workflow recognition \\
47 & M.6e.2 & MICCAI 2021 & PETRAW & Kinematic-based surgical workflow recognition \\
48 & M.6e.3 & MICCAI 2021 & PETRAW & Segmentation-based surgical workflow recognition \\
49 & M.6e.4 & MICCAI 2021 & PETRAW & Video and kinematic-based surgical workflow recognition \\
50 & M.6e.5 & MICCAI 2021 & PETRAW & Video, kinematic and   segmentation-based surgical workflow recognition \\
51 & M.6f.1 & MICCAI 2021 & SimSurgSkill & Surgical tool/needle detection \\
52 & M.6f.2 & MICCAI 2021 & SimSurgSkill & Skill Assessment \\
53 & M.7.1 & MICCAI 2021 & DiSCo & Quantitative connectivity estimation \\
54 & M.8.1 & MICCAI 2021 & FLARE21 & Abdominal Organ Segmentation in CT Images \\
55 & M.9.1 & MICCAI 2021 & FeTS & Federated Training \\
56 & M.9.2 & MICCAI 2021 & FeTS & Federated Evaluation \\
57 & M.10.1 & MICCAI 2021 & FeTA & Fetal Brain Tissue Segmentation \\
58 & M.11.1 & MICCAI 2021 & HECKTOR & Tumor segmentation \\
59 & M.11.2 & MICCAI 2021 & HECKTOR & Radiomics \\
60 & M.11.3 & MICCAI 2021 & HECKTOR & Radiomics with ground truth contour \\
61 & M.12.1 & MICCAI 2021 & LEARN2REG & Intra-patient multimodal abdominal MRI and CT registration \\
62 & M.12.2 & MICCAI 2021 & LEARN2REG & Intra-patient large deformation lung CT registration \\
63 & M.12.3 & MICCAI 2021 & LEARN2REG & Inter-patient large scale brain MRI registration \\
64 & M.13.1 & MICCAI 2021 & MOOD & Sample-level \\
65 & M.13.2 & MICCAI 2021 & MOOD & Pixel-level \\
66 & M.14.1 & MICCAI 2021 & MIDOG & Mitotic figure detection \\
67 & M.15.1 & MICCAI 2021 & M\&Ms-2 & Segmentation of the right ventricle (RV) in cardiac MRI \\
68 & M.16.1 & MICCAI 2021 & QUBIQ 2021 & Quantifying segmentation uncertainties \\
69 & M.17.1 & MICCAI 2021 & BraTS2021 & Segmentation of glioblastoma in mpMRI scans \\
70 & M.18.1 & MICCAI 2021 & SARAS-MESAD & Multi-domain static action detection \\
71 & M.19.1 & MICCAI 2021 & AutoImplant 2021 & Cranial implant design for diverse synthetic defects on aligned skulls \\
72 & M.19.2 & MICCAI 2021 & AutoImplant 2021 & Cranial implant design for real patient defects \\
73 & M.19.3 & MICCAI 2021 & AutoImplant 2021 & Improving the model generalization ability for cranial implant design \\
74 & M.20.1 & MICCAI 2021 & VALDO & Segmentation of enlarged PVS \\
75 & M.20.2 & MICCAI 2021 & VALDO & Segmentation of cerebral microbleeds \\
76 & M.20.3 & MICCAI 2021 & VALDO & Segmentation of lacunes \\
77 & M.21.1 & MICCAI 2021 & VWS & Vessel wall segmentation \\
78 & M.22.1 & MICCAI 2021 & FU-Seg & Foot Ulcer Segmentation \\
79 & M.23.1 & MICCAI 2021 & MSSEG-2 & New MS lesions segmentation \\
80 & M.24.1 & MICCAI 2021 & PAIP2021 & Detection of perineural invasion in three organ cancers \\* \bottomrule
\end{longtable}


\end{document}